\documentclass[conference, american, russian]{IEEEtran}
\usepackage[utf8]{inputenc}
\usepackage[russian, USenglish]{babel}
\usepackage[T1, T2A]{fontenc}
\usepackage{cite}
\ifCLASSINFOpdf
  \usepackage[pdftex]{graphicx}
\else
  \usepackage[dvips]{graphicx}
\fi
\usepackage{dblfloatfix}
\usepackage{graphbox}
\usepackage[nice]{nicefrac}
\usepackage[cmex10]{amsmath}
\usepackage{amssymb}
\usepackage{amsfonts}
\usepackage{algorithmic}
\usepackage{url}
\usepackage{textcomp}
\usepackage{xcolor}
\usepackage{comment}
\usepackage{gensymb}
\usepackage{siunitx}
\usepackage{tikz}
\usetikzlibrary{patterns}
\usepackage{tkz-euclide}
\usetikzlibrary{arrows}
\usetikzlibrary{positioning}
\usepackage[hypertexnames = false]{hyperref} 

\usepackage{enumitem}
\usepackage{rotating}
\usepackage{caption}
\usepackage{tablefootnote}
\usepackage[shortcuts]{extdash}

\onecolumn

\newcommand{\myref}[1]{\ref{#1}}

\newcommand{\QuotationMarks}[1]{``{#1}''}

\newcommand{\EuclidDistance}[2]{\operatorname{d}\!{\left( #1 , #2 \right)}}
\newcommand{\CurveDistance}[2]{\operatorname{\rho}\!{\left( #1 , #2 \right)}}

\newcommand\norm[1]{\left\lVert#1\right\rVert}

\DeclareMathOperator{\arcsinh}{arcsinh}

\DeclareMathOperator{\sech}{sech}

\newcommand{\Exp}{\operatorname{f}_{exp}\!}

\DeclareMathOperator{\sign}{sign}

\newcommand{\myvector}[1]{{\boldsymbol{\mathbf{ #1 }}}}

\newcommand{\myvectordef}[2]
{
  \begin{bmatrix}
    #1\\
    #2
  \end{bmatrix}
}

\begin{document}

\title{Reconstruction of Power Lines from Point Clouds}

\author
{
  \IEEEauthorblockN{Alexander Gribov\IEEEauthorrefmark{1} and Khalid Duri\IEEEauthorrefmark{2}}
  \IEEEauthorblockA
  {
    \textit{Esri}\\
    380 New York Street, Redlands, CA 92373-8100, USA \\
    \IEEEauthorrefmark{1}agribov@esri.com,
    \IEEEauthorrefmark{2}kduri@esri.com
  }
}

\maketitle

\begin{abstract}
  This paper proposes a novel solution for constructing line features modeling each catenary curve present within a~series of points representing multiple catenary curves. The solution can be applied to extract power lines from lidar point clouds, which can then be used in downstream applications like creating digital twin geospatial models and evaluating the encroachment of vegetation. This paper offers an example of how the results obtained by the proposed solution could be used to assess vegetation growth near transmission power lines based on freely available lidar data for the City of Utrecht, Netherlands~\cite{ReferenceCurrentHeightFileNetherlands}.
\end{abstract}
  

\section{Introduction}


This paper proposes a novel solution of constructing line features modeling each catenary curve that is present within a series of points representing multiple catenary curves. Utility companies can leverage this solution to extract power
lines from lidar surveys and identify specific attributes. The extracted power lines can be used to assess the locational accuracy of the power lines and determine whether the amount of sag (see calculation of sag in \cite{ReferenceCalculationOfSag}) indicates any need for maintenance of the support structure. Additionally, the extracted power lines can be incorporated into downstream operations like mitigating fire risk from vegetation encroachment. The power lines can help define the vegetation clearance zone surrounding them, as demonstrated later in this paper. The importance of maintaining power lines and their surrounding vegetation was underscored by the spate of fires in California caused by power lines (see Table~\ref{table:CaliforniaWildfires}). Of the fires related to power lines in~$ 2020 $, vegetation contact accounted for nearly $ 93 $~percent of the area damaged~\cite{ReferenceWildfireActivityStatistics2020}.

\begin{table} [htb]
  \caption[caption]
  {
    History of California wildfires caused by power lines.\cite{ReferenceWildfireActivityStatistics2020, ReferenceWildfireActivityStatistics2019, ReferenceWildfireActivityStatistics2018, ReferenceWildfireActivityStatistics2017}
    \label{table:CaliforniaWildfires}
  }
  \centering
  \begin{tabular}{ c | c | c | c | c | c }
    Year & Incident count & Incident count of acres burned $ \ge 10 $ & Acres burned & Dollar damage & \% of total burned area\\
    \hline
    $ 2020 $ & $ 335 $ & $ 31 $ & $  59,334 $ & $ \$     52,001,282 $\tablefootnote{Fires from undetermined causes resulted in $ \$ 2,121,798,631 $ in damages.} & $  4 $\%\\
    $ 2019 $ & $ 304 $ & $ 37 $ & $  83,729 $ & $ \$    388,843,293 $ & $ 65 $\%\\
    $ 2018 $ & $ 297 $ & $ 26 $ & $ 246,873 $ & $ \$  2,001,803,168 $ & $ 23 $\%\\
    $ 2017 $ & $ 408 $ & $ 65 $ & $ 249,501 $ & $ \$ 12,046,544,839 $ & $ 53 $\%\\
  \end{tabular}
\end{table}

Lidar provides a cost-effective solution for surveying the location of power lines and other objects of interest with great measures of accuracy. The points obtained from a lidar survey can be classified using a variety of techniques---from deterministic, rule-based operations and manual editing to deep learning. Once the points from power lines are classified, they can be processed with the proposed solution to produce a polyline representation for each power line. The resultant polylines can overcome gaps in the lidar survey and ensure the connectivity of each power line span.

The approach can be summarized as follows: 
\begin{enumerate} [label = \arabic*., ref = \arabic*]
  \item Use the minimum spanning tree clustering algorithm to connect points and cut them into sequences of \textit{combined} points (Section~\myref{sec:ClusteringCatenaryCurves}).
  \item Apply the dynamic programming approach to ensure the separation of each catenary curve (Section~\myref{DynamicProgrammingApproach}). The~dynamic programming approach uses a penalty function based on fitting the catenary curve to points (Section~\myref{sec:FittingCatenaryCurveToPoints}).
  \item Use the $ k $\=/mean clustering algorithm to improve partitioning. In $ k $\=/mean iterative steps, the update step refits catenary curves to the newly found clusters (Section~\myref{sec:FittingCatenaryCurveToPoints}) and the assignment step uses the algorithm to find the~closest points along the catenary curves (Section~\myref{sec:FindingTheClosestPointOnTheCatenaryCurve}).
\end{enumerate}

An example of the solution is shown in Section~\ref{sec:Example}.

\section{Clustering Catenary Curves\label{sec:ClusteringCatenaryCurves}}

Let's define a combined point as a nonempty set of points and a combined polyline as a list of combined points. A~combined polyline is considered small if it contains no more than the predetermined minimum number of combined points, and it is not small if it has more combined points than the predetermined maximum number of points. The prolongated form is discovered for all combined polylines whose points are in the range of the predetermined minimum and maximum number of points. This form is found by evaluating the ratio of the median axis to the largest axis of the ellipse constructed from the covariance matrix of the coordinates for each point in the combined polyline. This will be used to evaluate whether the predetermined threshold value has been exceeded.

The steps of the algorithm are the following:
\begin{enumerate} [label = \arabic*., ref = \arabic*]
  \item
    For the source points, apply the minimum spanning tree (MST) clustering algorithm using edges whose length does not exceed the user-specified threshold for the maximum allowed gap. See algorithms for the minimum spanning tree in~\cite{ReferenceBoruvka1, ReferenceBoruvka2, ReferenceOnTheShortestSpanningSubtreeOfAGraphAndTheTravelingSalesmanProblem, ReferenceShortestConnectionNetworksAndSomeGeneralizations}. Note the following: \textit{\QuotationMarks{Out of many available algorithms to solve MST, the Borůvka's algorithm is the basis of the fastest known algorithms}}~\cite{ReferenceBoruvka}.
  \item
    Assign each node of the network to an empty set of combined polylines.
  \item \label{enum:ProcessEndNodes}
    Store all nonsmall combined polylines for each end node with more than one nonsmall combined polyline, and discard all other combined polylines. Add the empty combined polyline to the neighboring node. To prevent merging combined polylines through this node, mark that empty combined polyline as not small.
    Perform the following for all other end nodes:
    \begin{enumerate} [label = \alph*., ref = \alph*]
      \item Find the longest combined polylines by the number of combined points, giving preference to nonsmall combined polylines.
      \item If there is only one longest combined polyline, take all points from other combined polylines, add these points and the node point as a combined point to the longest combined polyline, and add this modified combined polyline to the neighboring node.
      \item When two are the longest, construct a new combined polyline with a combined point containing all points from all combined polylines in the node and the node point, and add this combined polyline to the neighboring node.
    \end{enumerate}
  \item
    Delete all processed end nodes and go to step~\ref{enum:ProcessEndNodes} unless no end nodes remain.
  \item
    Note that at this stage, all nodes are isolated. Next, store all nonsmall combined polylines for each node with more than two nonsmall combined polylines. Otherwise, merge nonsmall combined polylines through the combined point with all points from other combined polylines and the node point.
  \item
    For all stored combined polylines, apply the division algorithm described in Section~\myref{DynamicProgrammingApproach}.
  \item
    Use the partitioning in the previous step as a starting solution for the $ k $\=/mean clustering algorithm~\cite{ReferenceKMean}. The~$ k $\=/mean clustering algorithm is then applied by alternating steps for fitting catenary curves to points and reallocating points to the closest catenary curves. Also, an essential intermediate step is merging similar catenary curves.
\end{enumerate}

This approach results in the identification of all points that belong to the same catenary curve. Consideration must be taken for the density of the points and expected length of the nominal catenary curve when determining appropriate values for the minimum and maximum number of points.

\section{Dividing a Combined Polyline\label{DynamicProgrammingApproach}}

The task is to find the optimal division of a combined polyline such that it minimizes the number of partitions and the penalty in each partition. In other words, each partition should have the best fit while encouraging the division of partitions when the fit improves significantly.

We found the following empirical penalties for each partition:
\begin{equation*}
  -\dfrac{\sum{\epsilon^2_i}}{2 \cdot n \cdot T^2}
  ,
\end{equation*}
where $ \epsilon_k $ is the deviation from the fitted catenary curve,
$ n $ is the number of points in the partition, and
$ T $ is the maximum deviation from the catenary curve specified by the user.
Fitting is performed without removing any points even if they violate the maximum deviation. Also, all points in a combined point cannot be divided.

And for the partitions not larger than the predetermined number of points, the penalty is 
\begin{equation*}
  \log{ \left( \dfrac{1}{2} \right) }
  ,
\end{equation*}
plus the total number of partitions times
$
  \log{ \left( \dfrac{1}{2} \right) }
$.

From a probabilistic point of view, this penalty formulation can be treated in the following way:
\begin{itemize}
  \item
    Each partition has a probability
    \begin{equation*}
      \Exp{ \left( \dfrac{\sum{\epsilon^2_i}}{2 \cdot n \cdot T^2} \right) }
      ,
    \end{equation*}
    where $ \Exp{} $ is the probability density function of the exponential distribution with an intensity equal to $ 1 $
    \begin{equation*}
      \Exp{ \left( x \right) }
      =
      \begin{cases}
        e^{-x}, & x \geq 0, \\
        0, & \text{otherwise.}
      \end{cases}
    \end{equation*}
    
  \item
    And for the partitions not larger than the user-specified number of points, the probability is $ \dfrac{1}{2} $. The~probability for each partition is $ \dfrac{1}{2} $, or in other words, the probability to have extra partition is $ \dfrac{1}{2} $ and the probability not to have any more partitions is $ \dfrac{1}{2} $. Therefore, the number of partitions follows the geometric distribution with probability equaling~$ \dfrac{1}{2} $.
\end{itemize}
The total probability is equal to the product of probabilities.

To find the optimal division, we use the dynamic programming approach described in \cite{PolylineGeneralizationCombinatorical}.

\section{Finding the Closest Point on the Catenary Curve\label{sec:FindingTheClosestPointOnTheCatenaryCurve}}

The catenary curve equation is
\begin{equation}
  y = c + a \cdot \cosh{\left( \dfrac{x - m}{a} \right)}
  .
  \label{equation:CatenaryCurve}
\end{equation}

\begin{figure*} [htb]
  \centering
  \includegraphics[width = 0.6\columnwidth, keepaspectratio]{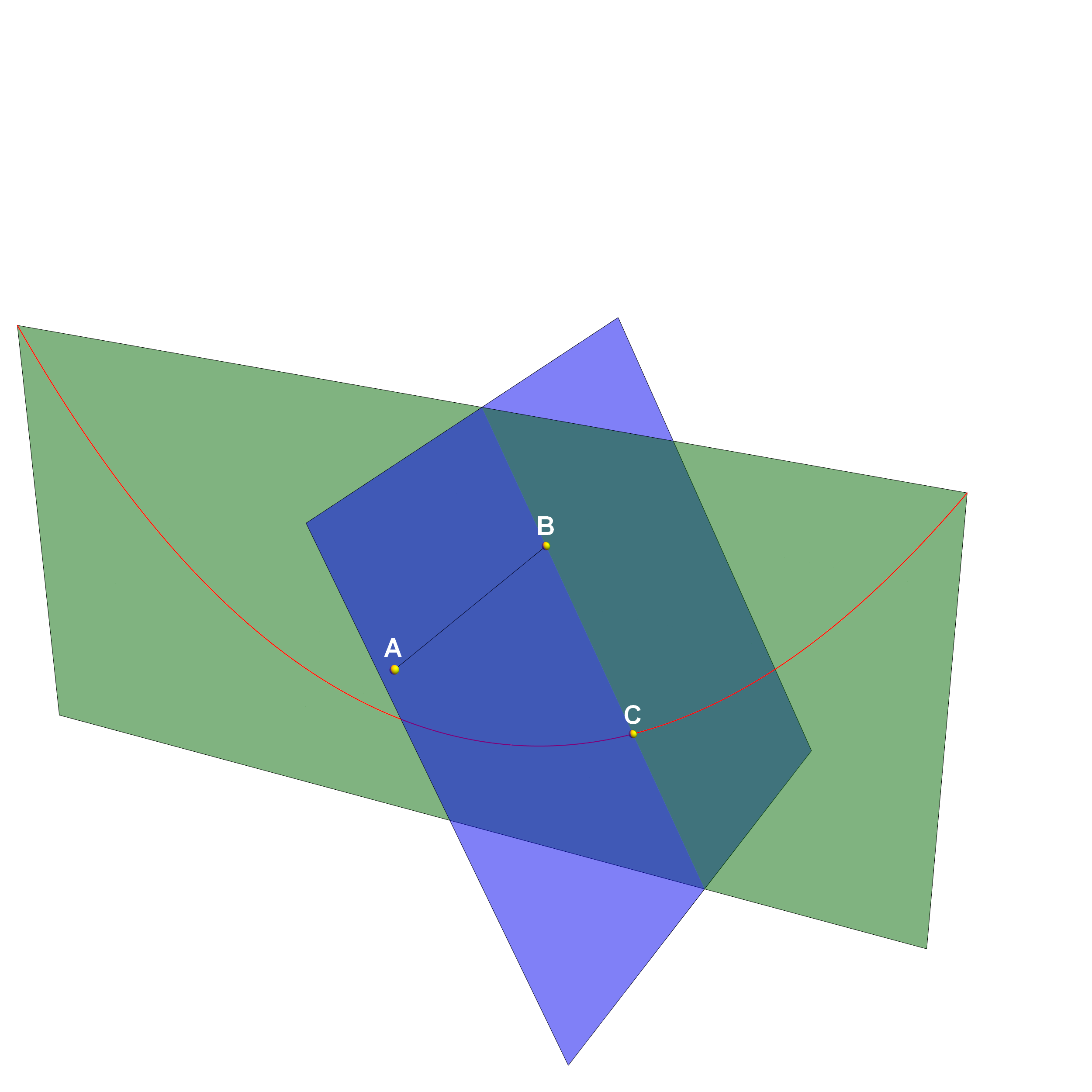}
  \caption
  {
    The catenary curve (red line) is defined in a plane (green). Point~$ B $ is a projection of point~$ A $ into the catenary curve's plane. Point~$ C $ is the point on the catenary curve closest to point~$ B $. The~tangent plane (blue) for the catenary curve is constructed for point~$ C $.
  }
  \label{fig:FindingClosestPointIn3D}
\end{figure*}

Finding the point~$ C $ on the catenary curve closest to some point~$ A $ in three-dimensional space is the same as finding the point~$ C $ on the catenary curve closest to the point~$ B $, where $ B $ is a projection of point~$ A $ onto the catenary curve's plane---see Figure~\ref{fig:FindingClosestPointIn3D}. Because point~$ B $ is a projection of point~$ A $ onto the catenary curve's plane and point~$ C $ is the closest point to point~$ B $, it follows that point~$ A $ is located on the tangent plane of the catenary curve constructed at point~$ C $. Therefore, finding the closest point in three-dimensional space can be done by projecting the point onto the catenary curve's plane, then determining the closest point on the catenary curve for the projected point.

To find the point closest to point $ \myvector{p} = \myvectordef{ x_p }{ y_p } $ on the catenary curve defined by \eqref{equation:CatenaryCurve}, we first apply the coordinate system's transformation:
\begin{align*}
  x' =& \dfrac{x - m}{a},\\
  y' =& \dfrac{y - c}{a}.
\end{align*}

After this transformation, the catenary curve \eqref{equation:CatenaryCurve} becomes
\begin{equation}
  y' = \cosh{\left( x' \right)},
  \label{equation:CanonicalCatenaryCurve}
\end{equation}
and point $ \myvector{p} $ transforms to $ \myvector{p}' = \myvectordef{ x_p' }{ y_p' } $, where
\begin{align*}
  x_p' =& \dfrac{x_p - m}{a},\\
  y_p' =& \dfrac{y_p - c}{a}.
\end{align*}

Because this transformation is translation and dilation, it follows that if the point $ \myvector{p}_c' = \myvectordef{ x_c' }{ y_c' } $ on the catenary curve~\eqref{equation:CanonicalCatenaryCurve} closest to the point is $ \myvector{p}' $, then $ \myvector{p}_c = \myvectordef{ x_c }{ y_c } $, where
\begin{align*}
  x_c =& a \cdot x_c' + m,\\
  y_c =& a \cdot y_c' + c
\end{align*}
is the location on the catenary curve \eqref{equation:CatenaryCurve} closest to the point $ \myvector{p} $.


Therefore, without loss of generality, it is sufficient to solve the task of finding the closest point on the catenary curve \eqref{equation:CatenaryCurve} by considering only the case when $ a = 1 $, $ m = 0 $, and $ c = 0 $; therefore, we will refer to the catenary curve as a function:
\begin{equation*}
  y = \cosh{\left( x \right)}
  .
\end{equation*}

\begin{figure*} [htb]
  \centering
  \begin{tikzpicture} [scale = 1.5]
    \tkzInit[xmin = -4.25, ymin = -0.25, xmax = 4.25, ymax = 5.25]

    \tkzGrid
    
    \begin{scope}
      \clip (-4.25, -0.25) rectangle (4.25, 5.25);
      \input{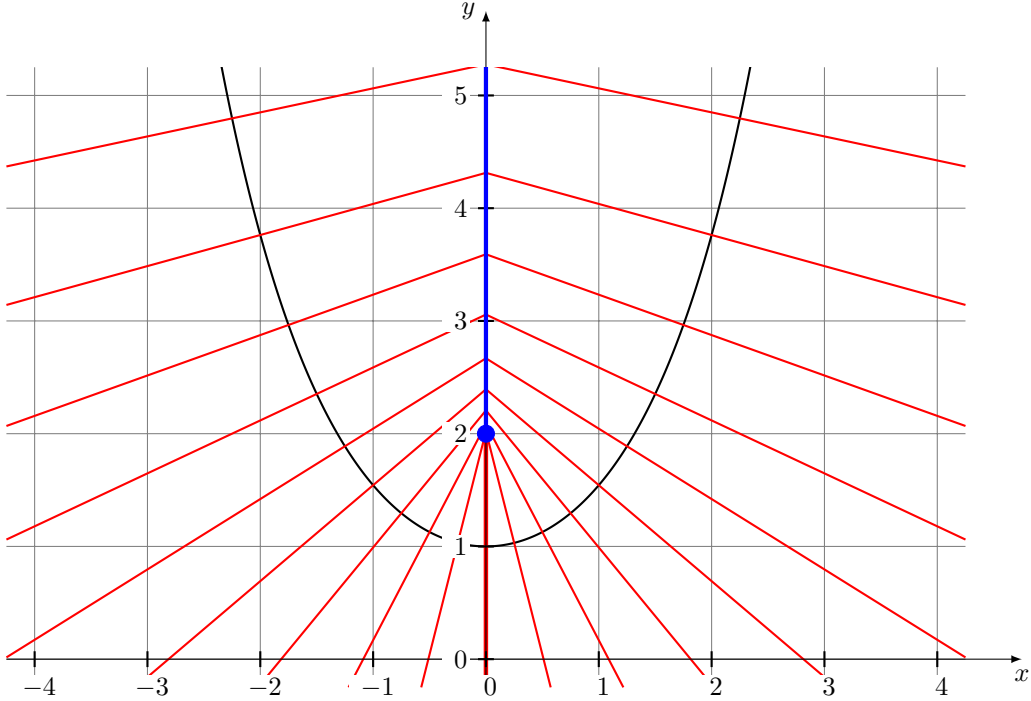}
        \draw [ultra thick, red] (-0.0, -0.25) -- (0.0, 2.0) -- (0.0, -0.25);
  \draw [thick, red] (-0.5737007319489158, -0.25) -- (0.0, 2.0210718907050733) -- (0.5737007319489158, -0.25);
  \draw [thick, red] (-1.2178744231953376, -0.25) -- (0.0, 2.0871433408738524) -- (1.2178744231953376, -0.25);
  \draw [thick, red] (-2.020218910531366, -0.25) -- (0.0, 2.2067406110970578) -- (2.020218910531366, -0.25);
  \draw [thick, red] (-3.1072305023344597, -0.25) -- (0.0, 2.393998763054565) -- (3.1072305023344597, -0.25);
  \draw [thick, red] (-4.25, 0.015670105206051055) -- (0.0, 2.668737948808918) -- (4.25, 0.015670105206051055);
  \draw [thick, red] (-4.25, 1.0608929036063794) -- (0.0, 3.056873276136084) -- (4.25, 1.0608929036063794);
  \draw [thick, red] (-4.25, 2.0682640232806415) -- (0.0, 3.5913353102413) -- (4.25, 2.0682640232806415);
  \draw [thick, red] (-4.25, 3.141824420347119) -- (0.0, 4.313636820627198) -- (4.25, 3.141824420347119);
  \draw [thick, red] (-4.25, 4.370234500053815) -- (0.0, 5.276192189667373) -- (4.25, 4.370234500053815);
  \draw [thick, red] (-4.25, 5.843043057417269) -- (0.0, 6.545498654301426) -- (4.25, 5.843043057417269);
  \draw [thick, red] (-4.25, 7.660709297149278) -- (0.0, 8.206325927869068) -- (4.25, 7.660709297149278);
  \draw [thick, red] (-4.25, 9.942885033691738) -- (0.0, 10.367126704784233) -- (4.25, 9.942885033691738);
  \draw [thick, red] (-4.25, 12.836891881972594) -- (0.0, 13.166968899266733) -- (4.25, 12.836891881972594);
  \draw [thick, red] (-4.25, 16.527487253550447) -- (0.0, 16.78439928608936) -- (4.25, 16.527487253550447);
  \draw [thick, red] (-4.25, 21.248769112607615) -- (0.0, 21.448780575597723) -- (4.25, 21.248769112607615);
  \draw [thick, red] (-4.25, 27.29907194343502) -- (0.0, 27.454807117319948) -- (4.25, 27.29907194343502);
  \draw [thick, red] (-4.25, 35.05983829029843) -- (0.0, 35.181108953268925) -- (4.25, 35.05983829029843);
  \draw [thick, red] (-4.25, 45.01967533236329) -- (0.0, 45.11411345752888) -- (4.25, 45.01967533236329);
  \draw [thick, red] (-4.25, 57.8051204540422) -- (0.0, 57.878665368240135) -- (4.25, 57.8051204540422);
  \draw [thick, red] (-4.25, 74.22005590416079) -- (0.0, 74.2773310539408) -- (4.25, 74.22005590416079);
    \end{scope}

    \tkzAxeXY

    \begin{scope}
      \clip (-4.25, -0.25) rectangle (4.25, 5.25);
      \draw [ultra thick, blue] (0, 2) -- (0, 100);
      \draw [blue, fill = blue] (0, 2) circle (0.075);
    \end{scope}
  \end{tikzpicture}
  \caption
  {
    The relationship between the points and the closest points on the catenary curve.
    The black line is a catenary curve; the red lines are normal lines to the catenary curve; the blue line is where two points on the catenary curve have the same distance; and the blue dot is an area of instability for finding the closest point on the catenary curve.
  }
  \label{fig:NormalsToCatenaryCurve}
\end{figure*}

The derivative at point $ x $ is $ \sinh{\left( x \right)} $, and it is strictly an increasing function. The normal to the catenary curve is $ \myvectordef{ \tanh{\left( x \right)} }{ -\sech{\left( x \right)} } $ (see Figure~\ref{fig:NormalsToCatenaryCurve}), and it will intersect the $ y $\=/axis at
\begin{equation}
  y = \cosh{\left( x \right)} + \dfrac{x}{\sinh{\left( x \right)}}
  .
  \label{equation:OYcrossing}
\end{equation}
Equation \eqref{equation:OYcrossing} is an even function, strictly increasing for positive $ x $, and strictly decreasing for negative $ x $; see proof in \ref{appendix:equation:crossing}. From these properties, it follows that the closest point is unique for any location except the line from point $ \myvectordef{ 0 }{ 2 } $ to $ \myvectordef{ 0 }{ +\inf } $, shown in blue. While the point $ \myvectordef{ 0 }{ 2 } $ has a unique closest point $ \myvectordef{ 0 }{ 1 } $, all other points on the blue line have two equally distant closest points. There is an area of numerical instability for finding the closest point for points close to the point $ \myvectordef{ 0 }{ 2 } $. In this paper, for all points on the line from point $ \myvectordef{ 0 }{ 2 } $ to $ \myvectordef{ 0 }{ +\inf } $ and around point $ \myvectordef{ 0 }{ 2 } $, point $ \myvectordef{ 0 }{ 1 } $ will be assigned as the closest.

Let's start by partitioning the area by normal lines from the catenary curve. Without loss of generality, it is sufficient to consider only half of the plane $ x > 0 $. Choosing $ x_i = k \cdot i $, $ i \in \mathbb{N}_0 $, construct lines passing through points $ \myvectordef{ x_i }{ \cosh{\left( x_i \right)} } $ with directions $ \myvectordef{ \tanh{\left( x_i \right)} }{ -\sech{\left( x_i \right)} } $. The example of partitioning the area for the case when $ k = \dfrac{1}{4} $ is shown in Figure~\ref{fig:NormalsToCatenaryCurve}.


\begin{figure*} [htb]
  \centering
  \begin{tikzpicture} [scale = 1.5]
    \tkzInit[xmin = 0, ymin = -0.25, xmax = 4.25, ymax = 6.25]

    \tkzGrid

    \begin{scope}
      \clip (0, -0.25) rectangle (4.25, 6.25);
      \input{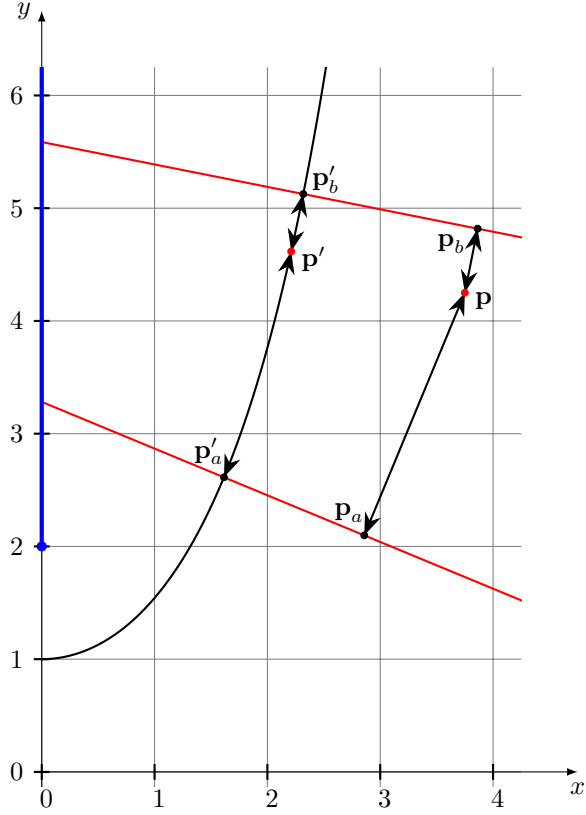}
    \end{scope}

    \begin{scope}
      \clip (-0.25, -0.25) rectangle (4.25, 6.25);
        \draw [thick, red] (0.0, 3.2820356489847624) -- (6.25, 0.6932008841529171);
  \draw [thick, red] (0.0, 5.5868671978462965) -- (6.25, 4.343664901723435);
    \end{scope}
    
    \draw [<->, {Stealth[length = 3mm]}-{Stealth[length = 3mm]}, thick, black] (3.75, 4.25) -- (2.8585981364499214, 2.0979655314927493);
    \draw (2.8585981364499214, 2.0979655314927493) node[xshift = -0.2cm, yshift = 0.3cm] {$ \myvector{p}_a $} [black, fill = black] circle (0.03);
    \draw [<->, {Stealth[length = 3mm]}-{Stealth[length = 3mm]}, thick, black] (3.75, 4.25) -- (3.8630725874027228, 4.8184542841265600);
    \draw (3.8630725874027228, 4.8184542841265600) node[xshift = -0.35cm, yshift = -0.2cm] {$ \myvector{p}_b $} [black, fill = black] circle (0.03);

    \draw [red, fill = red] (3.75, 4.25) node[black, xshift = 0.25cm, yshift = -0.1cm] {$ \myvector{p} $} circle (0.03);

    \draw [<-, {Stealth[length = 3mm]}-, thick, black] (1.6148909161730953, 2.6131259297527531) -- (1.6276041666666667, 2.644030436665486);
    \draw [->, -{Stealth[length = 3mm]}, thick, black] (2.2005208333333335, 4.570230357508402) -- (2.2107973526794366, 4.616300472864749);
    \draw [<-, {Stealth[length = 3mm]}-, thick, black] (2.2107973526794366, 4.616300472864749) -- (2.2135416666666665, 4.628685636560219);
    \draw [->, -{Stealth[length = 3mm]}, thick, black] (2.3177083333333335, 5.1254404030946805) -- (2.3177860101746115, 5.1258308954830124);
    \draw [black, fill = black] (1.6148909161730953, 2.6131259297527531) node[black, xshift = -0.2cm, yshift = 0.35cm] {$ \myvector{p}_a' $} circle (0.03);
    \draw [black, fill = black] (2.3177860101746115, 5.1258308954830124) node[xshift = 0.3cm, yshift = 0.2cm] {$ \myvector{p}_b' $} circle (0.03);
    \draw [red, fill = red] (2.2107973526794366, 4.616300472864749) node[black, xshift = 0.3cm, yshift = -0.1cm] {$ \myvector{p}' $} circle (0.03);

    \tkzAxeXY

    \begin{scope}
      \clip (-0.25, -0.25) rectangle (4.25, 6.25);
      \draw [ultra thick, blue] (0, 2) -- (0, 100);
      \draw [blue, fill = blue] (0, 2) circle (0.04);
    \end{scope}
  \end{tikzpicture}
  \caption
  {
    Example of partitioning by normal lines to the catenary curves (red lines) at points $ \myvector{p}_a' $ and $ \myvector{p}_b' $. $ \myvector{p}_a $ and $ \myvector{p}_b $ are the points on these normal lines closest to the point $ \myvector{p} $. $ \myvector{p}' $ approximates the point closest to the point $ \myvector{p} $ on the catenary curve. See the text for details.
  }
  \label{fig:NormalsToCatenaryCurve2}
\end{figure*}

Finding the partition is performed by a binary search. The bounds can be narrowed by finding the intersection of horizontal and vertical lines drawn from the point $ \myvector{p} $ with the catenary curve.

While this algorithm can be used for finding the catenary curve's closest point by subdividing on each iteration, many steps are needed to converge. However, if performance is not a concern, it can be the chosen algorithm for this task.

In the next step, using the found partition, the approximate location for the closest point is found by maintaining the same proportion of distances between the points and the region bounds. See Figure~\ref{fig:NormalsToCatenaryCurve2}, for the point $ \myvector{p} $, the distances are measured as the shortest distances to the region's bounds. For the point on the catenary curve $ \myvector{p}_c $, the distances are measured along the catenary curve to the region's bounds. Therefore,
\begin{equation*}
  \dfrac{\CurveDistance{ \myvector{p}_a' }{ \myvector{p}' }}{\CurveDistance{ \myvector{p}_a' }{ \myvector{p}_b' }}
  =
  \dfrac{\EuclidDistance{ \myvector{p}_a }{ \myvector{p} }}{\EuclidDistance{ \myvector{p}_a }{ \myvector{p} } + \EuclidDistance{ \myvector{p} }{ \myvector{p}_b }}
  ,
\end{equation*}
where $ \CurveDistance{ \myvector{p}_0 }{ \myvector{p}_1 } = \left| \sinh{\left( x_1 \right)} - \sinh{\left( x_0 \right)} \right| $ is the distance along the catenary curve between points $ \myvector{p}_0 = \myvectordef{ x_0 }{ \cosh{\left( x_0 \right)} } $ and $ \myvector{p}_1 = \myvectordef{ x_1 }{ \cosh{\left( x_1 \right)} } $; and $ \EuclidDistance{ \myvector{p}_0 }{ \myvector{p}_1 } $ is the Euclidean distance between points $ \myvector{p}_0 $ and $ \myvector{p}_1 $.\footnote{Because partitioning lines change directions, they are not parallel and, therefore, will intersect. Another way to find an approximate location of the closest point is by evaluating proportion from formed angles. This approach is less robust for larger values of $ x $.}

In the final step of finding the closest point, we will use an iterative algorithm. We consider two approximations: one by the osculating parabola, and another by the osculating circle.

\subsection{Approximation by the Osculating Parabola\label{sec:ApproximationByParabola}}

\begin{figure*} [htb]
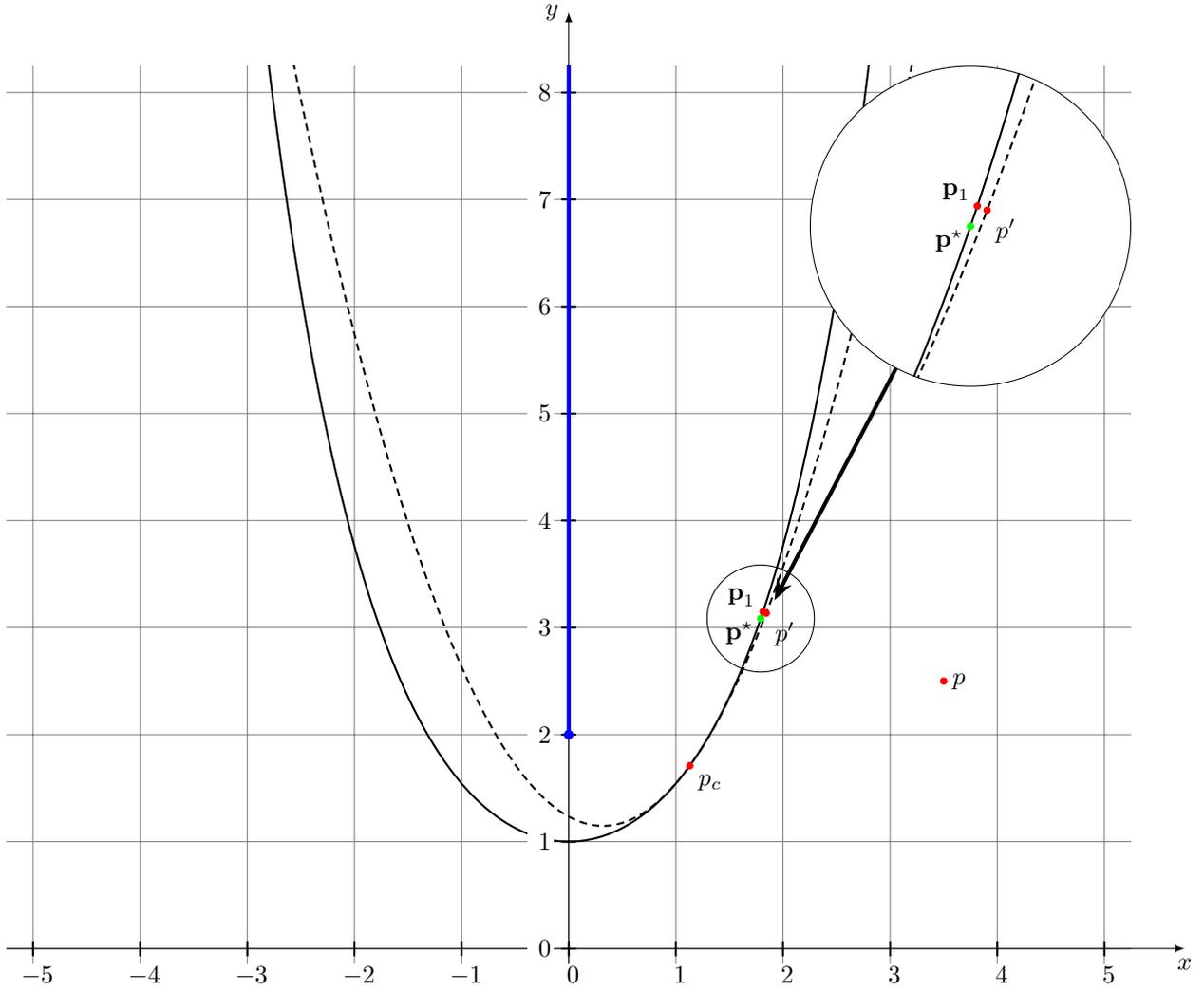

  \centering
  \begin{tikzpicture} [scale = 1.5]
    \tkzInit[xmin = -5.25, ymin = -0.25, xmax = 5.25, ymax = 8.25]
    
    \tkzGrid
    
    \begin{scope}
      \clip (-4.25, -0.25) rectangle (4.25, 8.25);
      \input{CatenaryCurveLine.tex}
      \input{CatenaryCurveParabola.tex}
    \end{scope}
    
    \draw [red, fill = red] (1.81372201460170124e+00, 3.14813956523205452e+00) node[black, xshift = -0.3cm, yshift = 0.2cm] {$ \myvector{p}_1 $} circle (0.03);

    \draw [green, fill = green] (1.79236504588954104e+00, 3.08510016406874144e+00) node[black, xshift = -0.3cm, yshift = -0.2cm] {$ \myvector{p}^\star $} circle (0.03);
    
    \tkzAxeXY
    
    \begin{scope}
      \clip (-5.25, -0.25) rectangle (5.25, 8.25);
      \draw [ultra thick, blue] (0, 2) -- (0, 100);
      \draw [blue, fill = blue] (0, 2) circle (0.04);
    \end{scope}
    
    \draw [black] (1.79236504588954104e+00, 3.08510016406874144e+00) circle (0.5);
    \draw [->, -{Stealth[length = 3mm]}, ultra thick, black] (3.75, 6.75) -- (1.91979, 3.25421);
    
    \node at (3.75, 6.75)
    {
      \begin{tikzpicture} [scale = 4.5]
        \begin{scope}
          \clip (1.79236504588954104e+00, 3.08510016406874144e+00) circle (0.5);
          \draw [thick, black, fill = white] (1.79236504588954104e+00, 3.08510016406874144e+00) circle (0.5);
          
          \input{CatenaryCurveLine2.tex}
          \input{CatenaryCurveParabola2.tex}
          
          \draw [red, fill = red] (1.81372201460170124e+00, 3.14813956523205452e+00) node[black, xshift = -0.3cm, yshift = 0.2cm] {$ \myvector{p}_1 $} circle (0.01);

          \draw [green, fill = green] (1.79236504588954104e+00, 3.08510016406874144e+00) node[black, xshift = -0.3cm, yshift = -0.2cm] {$ \myvector{p}^\star $} circle (0.01);
        \end{scope}
      \end{tikzpicture}
    };
  \end{tikzpicture}
  \caption
  {
    Approximation of a catenary curve (black line) by a parabola (dashed line) at the point $ \myvector{p}_c $. This~represents one step of the algorithm for finding the closest point on the catenary curve for the point $ \myvector{p} $ as point $ \myvector{p}_1 $. The point $ \myvector{p}' $ is the projection of the point $ \myvector{p} $ onto the parabola. The point $ \myvector{p}_1 $ is found to have the same distance along the catenary curve as the point $ \myvector{p}' $ along the parabola from the point $ \myvector{p}_c $. $ \myvector{p}^\star $ is the point on the catenary curve closest to the point~$ \myvector{p} $.
  }
  \label{fig:NormalsToCatenaryParabola}
\end{figure*}

In point $ \myvector{p}_c = \myvectordef{ x_c }{ y_c } $, where $ y_c = \cosh{\left( x_c \right)} $, we will approximate the catenary curve as an osculating parabola; see Figure~\ref{fig:NormalsToCatenaryParabola}. The equation for the osculating parabola is
$
  y{\left( x \right)} = \cosh{\left( x_c \right)} + \sinh{\left( x_c \right)} \cdot {\left( x - x_c \right)} + \frac{1}{2} \cdot \cosh{\left( x_c \right)} \cdot {\left( x - x_c \right)}^2
$.
The next step is to find the projection of point $ \myvector{p} $ onto the osculating parabola. The projected point $ \myvector{p}' $ will satisfy
\begin{equation*}
  \left( \myvector{p} - \myvector{p}' \right)
  \perp
  \myvector{d}'
  ,
\end{equation*}
where
\begin{equation*}
  \myvector{p}'
  =
  \begin{bmatrix}
    x_c + \Delta{x}\\
    \cosh{\left( x_c \right)} + \sinh{\left( x_c \right)} \cdot \Delta{x} + \frac{1}{2} \cdot \cosh{\left( x_c \right)} \cdot {\Delta{x}}^2
  \end{bmatrix}
\end{equation*}
and
\begin{equation*}
  \myvector{d}'
  =
  \begin{bmatrix}
    1\\
    \sinh{\left( x_c \right)} + \cosh{\left( x_c \right)} \cdot \Delta{x}
  \end{bmatrix}
\end{equation*}
is the tangent vector to the parabola at point $ \myvector{p}' $.

Measuring the distance from point $ \myvector{p}_c $ to $ \myvector{p}' $ along the parabola (see \ref{appendix:length_along_parabola}) and finding point $ \myvector{p}_1 $ on the same distance from point $ \myvector{p}_c $ measured on the catenary curve will produce the next approximation of the closest point on the catenary curve.

Two vectors are orthogonal if\textemdash and only if\textemdash their scalar product is zero; therefore,
\begin{equation*}
  \begin{aligned}
    &-\dfrac{1}{2} \cdot \cosh^2{\left( x_c \right)} \cdot {\Delta{x}}^3
    -\dfrac{3}{2} \cdot \cosh{\left( x_c \right)} \cdot \sinh{\left( x_c \right)} \cdot {\Delta{x}}^2
    -\left( \cosh^2{\left( x_c \right)} - \left( y_p - y_c \right) \cdot \cosh{\left( x_c \right)} \right) \cdot \Delta{x}+\\
    &+\left( \left( x_p - x_c \right) + \left( y_p - y_c \right) \cdot \sinh{\left( x_c \right)} \right)
    =
    0
    .
  \end{aligned}
\end{equation*}
Dividing this equation by $ -\frac{1}{2} \cdot \cosh^2{\left( x_c \right)} $ produces
\begin{equation}
  \begin{aligned}
    &{\Delta{x}}^3
    +3 \cdot \tanh{\left( x_c \right)} \cdot {\Delta{x}}^2
    +2 \cdot \left( 2 - y_p \cdot \sech{\left( x_c \right)} \right) \cdot \Delta{x}+\\
    &+2 \cdot \left( \left( x_c - x_p \right) \cdot \sech^2{\left( x_c \right)} + \left( 1 - y_p \cdot \sech{\left( x_c \right)} \right) \cdot \tanh{\left( x_c \right)} \right)
    =
    0
    .
  \end{aligned}
  \label{eq:CubicEquationForOsculatingParabola}
\end{equation}
Cubic equations with real coefficients can only have either one or three real roots. Without loss of generality, it is assumed that $ x_p > 0 $. It can be shown that if there are three roots, the closest point on the parabola will correspond to the largest root; see \myref{RootsForCubicEquationForOsculatingParabola} for details.

Taking the solution of the cubic equation from \cite[chapter~5.6]{NumericalRecipes}, it follows that
\begin{equation*}
  {\Delta{x}}^3 + 3 \cdot a' \cdot {\Delta{x}}^2 + 2 \cdot b' \cdot \Delta{x} + 2 \cdot c' = 0
  ,
\end{equation*}
where
\begin{align*}
  a' &= \tanh{\left( x_c \right)},\\
  b' &= \left( 2 - y_p \cdot \sech{\left( x_c \right)} \right), \text{ and}\\
  c' &= \left( x_c - x_p \right) \cdot \sech^2{\left( x_c \right)} + \left( 1 - y_p \cdot \sech{\left( x_c \right)} \right) \cdot \tanh{\left( x_c \right)}.
\end{align*}

\begin{equation*}
  \begin{aligned}
    Q &= a'^2 - \dfrac{2}{3} b',\\
    R &= a'^3 - a' \cdot b' + \dfrac{1}{2}c'.
  \end{aligned}
\end{equation*}

If $ R^2 < Q^3 $ the largest root is
\begin{equation*}
  \Delta{x} = 2 \cdot \sqrt{Q} \cdot \cos{\left( \dfrac{\Theta}{3} \right)} - a',
\end{equation*}
where
$
  \Theta = \arccos{\left( -\dfrac{R}{Q^{\frac{3}{2}}} \right)}
$;
otherwise,
\begin{equation*}
  \Delta{x} = \left( A + B \right) - a',
\end{equation*}
where
\begin{align*}
  A &= -\sign{\left( R \right)} \cdot {\left( \left| R \right| + \sqrt{R^2 - Q^3} \right)}^{\frac{1}{3}},\\
  B &=
  \begin{cases}
    \dfrac{Q}{A}, & \text{if } A \neq 0;\\
    0, & \text{otherwise}.
  \end{cases}
\end{align*}

Due to a round off error, the found correction $ \Delta{x} $ can be significantly off, requiring some iterative algorithm to improve precision of the solution for the cubic equation; see \cite[chapter~9.5]{NumericalRecipes}. Applying Laguerre's method \cite[chapter~9.5.3]{NumericalRecipes} produces good results.

\subsection{Approximation by the Osculating Circle}

\begin{figure*} [htb]
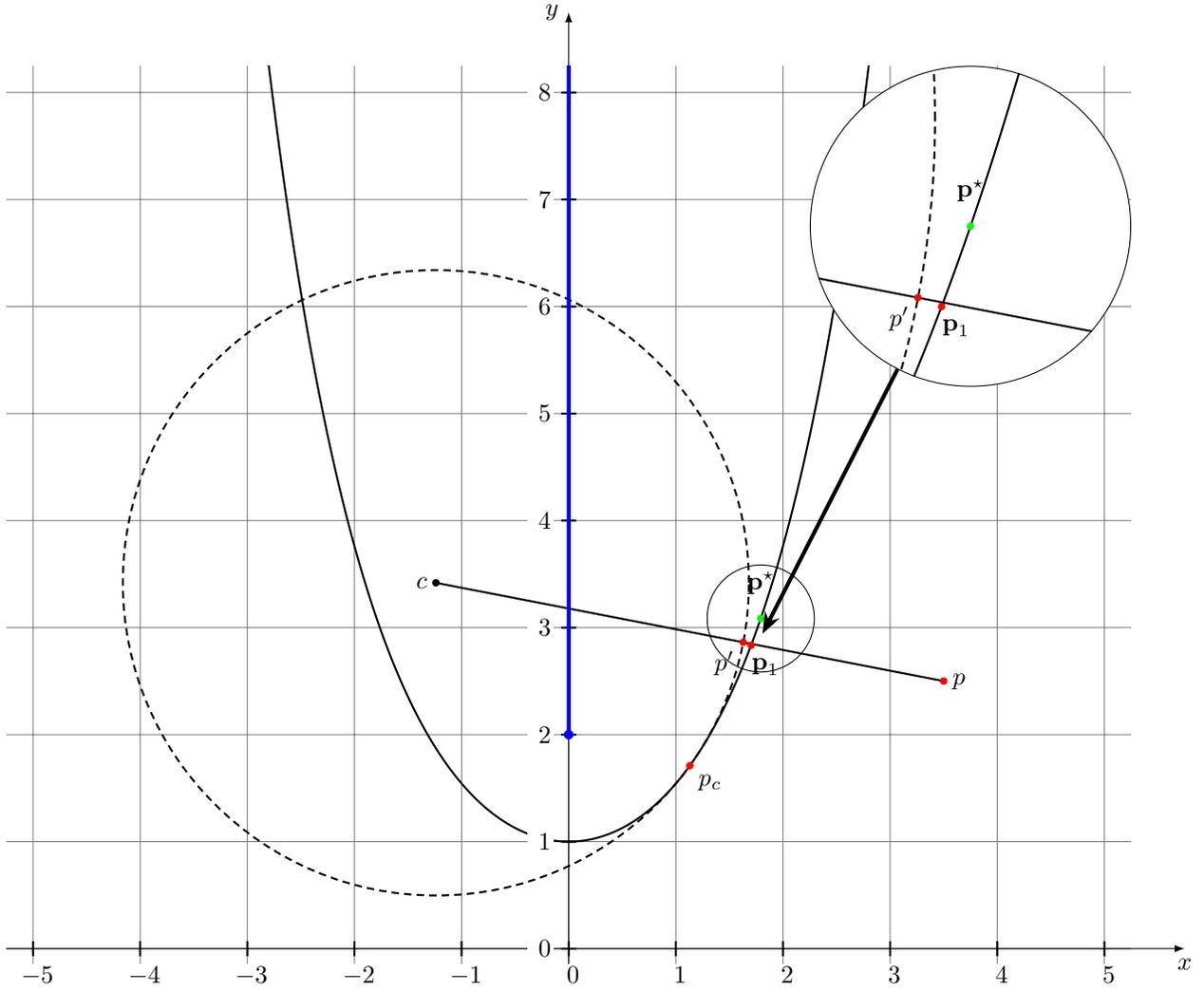

  \centering
  \begin{tikzpicture} [scale = 1.5]
    \tkzInit[xmin = -5.25, ymin = -0.25, xmax = 5.25, ymax = 8.25]
    
    \tkzGrid
    
    \begin{scope}
      \clip (-4.25, -0.25) rectangle (4.25, 8.25);
      \input{CatenaryCurveLine.tex}
          \draw [thick, densely dashed, black]  (-1.239684670502403, 3.4186897565469643) circle (2.9218599128797855);
    \draw [thick, black] (-1.239684670502403, 3.4186897565469643) -- (3.5, 2.5);
    \draw [red, fill = red] (1.13, 1.7093448782734821) node[black, below right] {$ p_c $} circle (0.03);
    \draw [red, fill = red] (3.5, 2.5) node[black, right] {$ p $} circle (0.03);
    \draw [red, fill = red] (1.628788085556792, 2.8626956934131633) node[black, below left] {$ p' $} circle (0.03);
    \draw [black, fill = black] (-1.239684670502403, 3.4186897565469643) node[black, left] {$ c $} circle (0.03);
    \end{scope}
    
    \draw [red, fill = red] (1.70252729182797236e+00, 2.83501078149264352e+00) node[black, xshift = 0.2cm, yshift = -0.3cm] {$ \myvector{p}_1 $} circle (0.03);
    
    \draw [green, fill = green] (1.79236504588954104e+00, 3.08510016406874144e+00) node[black, xshift = 0cm, yshift = 0.5cm] {$ \myvector{p}^\star $} circle (0.03);
    
    \tkzAxeXY
    
    \begin{scope}
      \clip (-5.25, -0.25) rectangle (5.25, 8.25);
      \draw [ultra thick, blue] (0, 2) -- (0, 100);
      \draw [blue, fill = blue] (0, 2) circle (0.04);
    \end{scope}
    
    \draw [black] (1.79236504588954104e+00, 3.08510016406874144e+00) circle (0.5);
    \draw [->, -{Stealth[length = 3mm]}, ultra thick, black] (3.75, 6.75) -- (1.80859, 2.94108);
    
    \node at (3.75, 6.75)
    {
      \begin{tikzpicture} [scale = 4.5]
        \begin{scope}
          \clip (1.79236504588954104e+00, 3.08510016406874144e+00) circle (0.5);
          \draw [thick, black, fill = white] (1.79236504588954104e+00, 3.08510016406874144e+00) circle (0.5);
          
          \input{CatenaryCurveLine2.tex}
              \draw [thick, densely dashed, black]  (-1.239684670502403, 3.4186897565469643) circle (2.9218599128797855);
    \draw [thick, black] (-1.239684670502403, 3.4186897565469643) -- (3.5, 2.5);
    \draw [red, fill = red] (1.13, 1.7093448782734821) node[black, below right] {$ p_c $} circle (0.01);
    \draw [red, fill = red] (3.5, 2.5) node[black, right] {$ p $} circle (0.01);
    \draw [red, fill = red] (1.628788085556792, 2.8626956934131633) node[black, below left] {$ p' $} circle (0.01);
    \draw [black, fill = black] (-1.239684670502403, 3.4186897565469643) node[black, left] {$ c $} circle (0.01);
          
          \draw [red, fill = red] (1.70252729182797236e+00, 2.83501078149264352e+00) node[black, xshift = 0.2cm, yshift = -0.3cm] {$ \myvector{p}_1 $} circle (0.01);
          
          \draw [green, fill = green] (1.79236504588954104e+00, 3.08510016406874144e+00) node[black, xshift = 0cm, yshift = 0.5cm] {$ \myvector{p}^\star $} circle (0.01);
        \end{scope}
      \end{tikzpicture}
    };
  \end{tikzpicture}
  \caption
  {
    Approximation of a catenary curve (black line) by an osculating circle (dashed line) at the point $ \myvector{p}_c $. This~represents one step of the algorithm for finding the closest point on the catenary curve for the point $ \myvector{p} $ as point~$ \myvector{p}_1 $. The point $ \myvector{p}' $ is the projection of the point $ \myvector{p} $ onto the osculating circle. The point $ \myvector{p}_1 $ is found to have the same distance along the catenary curve as the point $ \myvector{p}' $ along the circle from the point $ \myvector{p}_c $. The point $ \myvector{p}^\star $ is the location on the catenary curve closest to the point $ \myvector{p} $.
  }
  \label{fig:NormalsToCatenaryCircle}
\end{figure*}

In point $ \myvector{p}_c = \myvectordef{ x_c }{ y_c } $, where $ y_c = \cosh{\left( x_c \right)} $, we will approximate the catenary curve as an osculating circle; see Figure~\ref{fig:NormalsToCatenaryCircle}. The osculating circle will have the radius $ \cosh^2{\left( x_c \right)} $ and center at $ \myvectordef{ x_c - \sinh{\left( x_c \right)} \cdot \cosh{\left( x_c \right)} }{ y_c + \cosh{\left( x_c \right)} } $. \footnote{Notice that the $ x $\=/coordinate of the osculating circle center is of the opposite sign of $ x_c $ unless $ x_c $ is equal to $ 0 $; and the~$ y $\=/coordinate is always twice the $ y_c $.} Point $ \myvector{p}' $ is the projection of point $ \myvector{p} $ onto the osculating circle. Measuring the distance on the circle from point~$ \myvector{p}_c $ to $ \myvector{p}' $ and finding point $ \myvector{p}_1 $ at the same distance from point $ \myvector{p}_c $ measured on the catenary curve will produce the next approximation of the closest point on the catenary curve.

Let $ \myvector{p}_c = \myvectordef{ x_c }{ \cosh{\left( x_c \right)} } $ be the current approximation of the point closest to the point $ \myvector{p} $.
The normal vector to the catenary curve at the point $ \myvector{p}_c $ will be $ v_n =  \left( x_n, y_n \right) = \left( \tanh{\left( x_c \right)}, -\sech{\left( x_c \right)} \right) $.
Rotate vector $ \myvector{p} - \myvector{p}_c $ in the opposite direction of vector $ v_n $. This should rotate vector $ \myvector{v}_n $ to $ \myvectordef{ 1 }{ 0 } $.
\begin{equation*}
  \myvector{v}_r
  =
  \begin{bmatrix}
    x_r\\
    y_r
  \end{bmatrix}
  =
  \begin{bmatrix}
    x_n & y_n\\
    -y_n & x_n
  \end{bmatrix}
  \cdot
  \left( p - p_c \right)
  .
\end{equation*}

The radius of the osculating circle is
\begin{equation*}
  r = \cosh^2{\left( x_c \right)}
  .
\end{equation*}

The chordal distance between $ \myvector{p}_c $ and $ \myvector{p}' $ is
\begin{equation*}
  l_c
  =
  \sqrt
  {
    \left(
      r
      \cdot
      \left(
      \dfrac
      {
        1
      }
      {
        \sqrt
        {
          \left(
          1
          +
          {\left(
            \dfrac{y_r}{r + x_r}
            \right)}^2
          \right)
        }
      }
      -
      1
      \right)
    \right)^2
    +
    \left(
      \dfrac
      {
        y_r
      }
      {
        \sqrt
        {
          {\left( 1 + \dfrac{x_r}{r} \right)}^2
          +
          {\left( \dfrac{y_r}{r} \right)}^2
        }
      }
    \right)^2
  }
  .
\end{equation*}
In this equation, dividing first by $ r + x_r $ and $ r $ improves the stability of the numeric evaluation.

\begin{equation*}
  l
  =
  \begin{cases}
    2 \cdot r \cdot \arcsin{\left( \dfrac{l_c}{2 \cdot r} \right)}, &  \text{if } \num{2.14911933289082095e-08} < \dfrac{l_c}{2 \cdot r};\\
    l_c, & \text{otherwise}.
  \end{cases}
\end{equation*}
In this equation, to prevent underflow, the constant $ \num{2.14911933289082095e-08} $ was used to skip evaluation of $ \arcsin $ for small values. This constant is the largest number for which $ \arcsin $ does not modify the number; however, it depends on the implementation of the $ \arcsin $ function. In the case where $ r $ overflows, $ l = \left| y_r \right| $.

Taking the distance $ l $ in the direction of $ y_r $ from point $ \myvector{p}_c $ along the catenary curve will yield point $ \myvector{p}_1 $, which is the next approximation for the point on the catenary curve closest to point $ \myvector{p} $.

\subsection{Choice for the Approximation Algorithm}

Both approximation algorithms work well and can be used to find the closest point on the catenary curve. While approximating by the osculating parabola produces a more precise solution than using the osculating circle, the latter produces a faster solution. About three iterations are sufficient to reach a relative precision of $ 10^{-6} $.

\section{Fitting a Catenary Curve to Points\label{sec:FittingCatenaryCurveToPoints}}

In the previous section, the task of finding the closest point on the catenary curve was solved. This section will describe the algorithm for fitting a catenary curve to a set of points using a nonlinear least squares algorithm where the parameters of the curve are optimized to minimize the sum of minimum distances between the source points and the curve.

Let's define a signed distance from a point $ \myvector{p} = \myvectordef{ x_p }{ y_p } $ to the catenary curve as a scalar product of the normal to the catenary curve at the point $ \myvector{p}_c $ and $ \overrightarrow{ \myvector{p}_c, \myvector{p} } $,
\begin{equation*}
  F{\left( c, a, m | \myvector{p}, \myvector{p}_c \right)}
  =
  \left( \tanh{\left( \dfrac{x_c - m}{a} \right)}, -\sech{\left( \dfrac{x_c - m}{a} \right)} \right)
  \cdot
  \left(
    x_p - x_c,
    y_p - (c + a \cdot \cosh{\left( \dfrac{x_c - m}{a} \right)}
  \right)
  ,
\end{equation*}
where
$ \myvector{p}_c = \myvectordef{ x_c }{ y_c } $, $ y_c = c + a \cdot \cosh{\left( \dfrac{x_c - m}{a} \right)} $, is the point on the catenary curve closest to $ \myvector{p} $. Point~$ \myvector{p}_c $ is found using the algorithm described in Section~\myref{sec:FindingTheClosestPointOnTheCatenaryCurve}.

This equation will be used in the trust region algorithm \cite{ReferenceBookTrustRegionMethods, ReferenceLeastSquareFittingKennethLevenberg0, ReferenceLeastSquareFittingDonaldWMarquardt1, ReferenceLeastSquareFittingAnanthRanganathan2, ReferenceLeastSquareFittingHenriPGavin3} to find optimal parameters $ c $, $ a $, and $ m $ by minimizing the sum of $ \norm{F{\left( c, a, m | \myvector{p}_i, \myvector{p}_{c, i} \right)}}^2 $ over a set of points $ \myvector{p}_i $, where $ \myvector{p}_{c, i} $ is the point on the catenary curve closest to the point~$ \myvector{p}_i $. The algorithm requires an evaluation of the function and its first derivatives,
\begin{equation*}
  \begin{aligned}
    &
    F{\left( c, a, m | p, p_c \right)}
    =
    a + \sech{\left( \frac{x_c-m}{a} \right)} \cdot \left( \left(x_p - x_c\right) \cdot \sinh{\left( \frac{x_c-m}{a} \right)} - \left( y_p - c \right) \right),\\
    &
    F_c'{\left( c, a, m | p, p_c \right)}
    =
    \sech{\left( \frac{x_c - m}{a} \right)},\\
    &
    F_a'{\left( c, a, m | p, p_c \right)}
    =
    \frac{x_c - m}{a} \cdot F_m'{\left( c, a, m | p, p_c \right)} + 1,\\
    &
    F_m'{\left( c, a, m | p, p_c \right)}
    =
    -\frac{\sech^2{\left(\frac{x_c - m}{a}\right)} \cdot \left( \left( y_p - c \right) \cdot \sinh \left(\frac{x_c - m}{a}\right) + \left( x_p - x_c \right) \right)}{a}.
  \end{aligned}
\end{equation*}
While the $ x $\=/coordinate of the closest point location is not moving when catenary curve parameters change, the scalar product to the normal keeps only the orthogonal distance to the linear approximation at location $ \myvector{p}_c $; see examples in Figure~\ref{fig:LinearApproximationOfDistance}. This approach is essentially a linear approximation compared to the distance calculated using the fixed point on the catenary curve. 

\begin{figure*} [htb]
  \centering
  \begin{tikzpicture} [scale = 2]
    \tkzInit[xmin = -0.25, ymin = -0.25, xmax = 1.25, ymax = 2.25]
    
    \tkzGrid
    
    \begin{scope}
      \clip (-0.25, -0.25) rectangle (1.25, 2.25);
      \input{CatenaryCurveLine.tex}
      \input{ApproximationOfDistance2.tex}
    \end{scope}
   
    \tkzAxeXY
  \end{tikzpicture}
  \captionsetup{singlelinecheck=off}
  \caption
  {
    Example of approximation of the shortest distance from the point to the catenary curve when the catenary curve is moving. The distance from the point $ \myvector{p} $ to the catenary curve (thin black line) is approximated as the shortest distance $ \protect\overrightarrow{ \myvector{p}, \myvector{q} } $ (dotted line) to the tangent line (dashed line) constructed at the point on the catenary curve $ \myvector{p}_c' $. That distance can be found as the absolute value of the scalar product of the normal to the catenary curve at point~$ \myvector{p}_c' $ and $ \protect\overrightarrow{ \myvector{p}, \myvector{p}_c' } $. The point $ \myvector{p}_c' $ on the catenary curve is found to have the same $ x $\=/coordinate as the point $ \myvector{p}_c $ on the source catenary curve (thick black line) closest to the point $ \myvector{p} $.
  }
  \label{fig:LinearApproximationOfDistance}
\end{figure*}

The domain of parameter $ a $ is $ \left( 0; +\inf \right) $. To apply the trust region algorithm without constraint, we will make the following substitution: $ a = e^{a_e} $. The first derivative for $ a_e $ is
\begin{equation*}
  F_{a_e}'{\left( c, e^{a_e}, m | \myvector{p}, \myvector{p}_c \right)} = F_a'{\left( c, e^{a_e}, m | \myvector{p}, \myvector{p}_c \right)} \cdot e^{a_e}.
\end{equation*}

The steps of the algorithm are the following:
\begin{enumerate} [label = \arabic*., ref = \arabic*]
  \item \label{enum:FittingPlane} Find the vertical plane that best fits the original points in terms of the least squares distance from the plane.
  
  \item Define a coordinate system on the fitted plane with the $ x $\=/axis being horizontal and the $ y $\=/axis being vertical. The~$ x $\=/axis can be found by projecting the normal vector to the horizontal plane, normalizing it, and rotating clockwise by $ 90 $ degrees. The $ y $\=/axis can be found by the vector product of the normal and the found $ x $\=/axis direction.
  
  \item Project source points onto the plane. Let projected points be $ \myvector{p}_i = \myvectordef{ x_i }{ y_i } $, where $ i = \overline{1..n} $, and $ n $~is the number of points.

  \item Fit the parabola to the points.

  \item Check if the parabola is close to the straight line or concave. If it is, the catenary curve does not represent the points.
  
  \item Find the center of mass for the projected points.
  
  \item Find the closest point on the parabola for the found center. It is done through solving a cubic equation; see~\cite{NumericalRecipes} and Section~\myref{sec:ApproximationByParabola}.
  
  \item \label{enum:FittingCatenaryCurveToParabola} Find catenary curve parameters $ a_0 $ and $ m_0 $ by matching the first and second derivatives of the catenary curve at the closest point on the parabola. The parameter $ c_0 $ is found to minimize the sum of squared deviations between the catenary curve and the data (average difference between the points and the catenary curve). This will produce initial estimates for catenary curve parameters $ c_0 $, $ a_0 $, and $ m_0 $.
  
  \item Apply the trust region algorithm.\footnote{Note that the result of this algorithm does not depend on the distance of the points to the fitted catenary curve plane (projection plane) because the squared distance to any point on the projection plane is the sum of the squared distance to the projection plane and the squared distance from the projected point to that point. However, squared distances from the points to the projection plane are fixed.}
\end{enumerate}

We have noticed that some power lines captured in lidar surveys are slightly rotated by the wind. In such cases, in step~\ref{enum:FittingPlane}, the optimal plane's orientation can be found to account for the influence of the wind in lieu of fitting to a~vertical plane. We assume that the wind blows in one direction with consistent strength.\footnote{Strong or inconsistent winds acting on a long power line can produce irregular distortions of its shape that create noncatenary curves. No examples of this were discovered when surveying available data for power lines and therefore were not considered in this paper.}

This algorithm is applied to real data that often contains outliers introduced by the following: 
\begin{itemize}
  \item Sensor noise producing improper points during the lidar survey
  \item Misclassified points that do not belong to any catenary curve
  \item Points from a different catenary curve during the clustering process
\end{itemize}
For these reasons, it became necessary to provide a user-specified maximum deviation threshold from the catenary curve that can be tailored to the data being processed. To make this algorithm robust to outliers, points exceeding this threshold need to be removed, and the algorithm needs to be reapplied to the remaining points.

Note that fitting a plane in step~\ref{enum:FittingPlane} does not find the optimal plane for the catenary curve but rather a close approximation. Finding an optimal plane requires simultaneous optimization of parameters for the catenary curve and the plane.

\section{Example\label{sec:Example}}

\begin{sidewaysfigure} [p]
  \centering
  \begin{tabular} {l r}
    \begin{tikzpicture}
      \node at (0, 0) [above right] {\includegraphics[trim = {{300 px} 0 0 0}, clip, width = 0.4725\columnwidth, keepaspectratio]{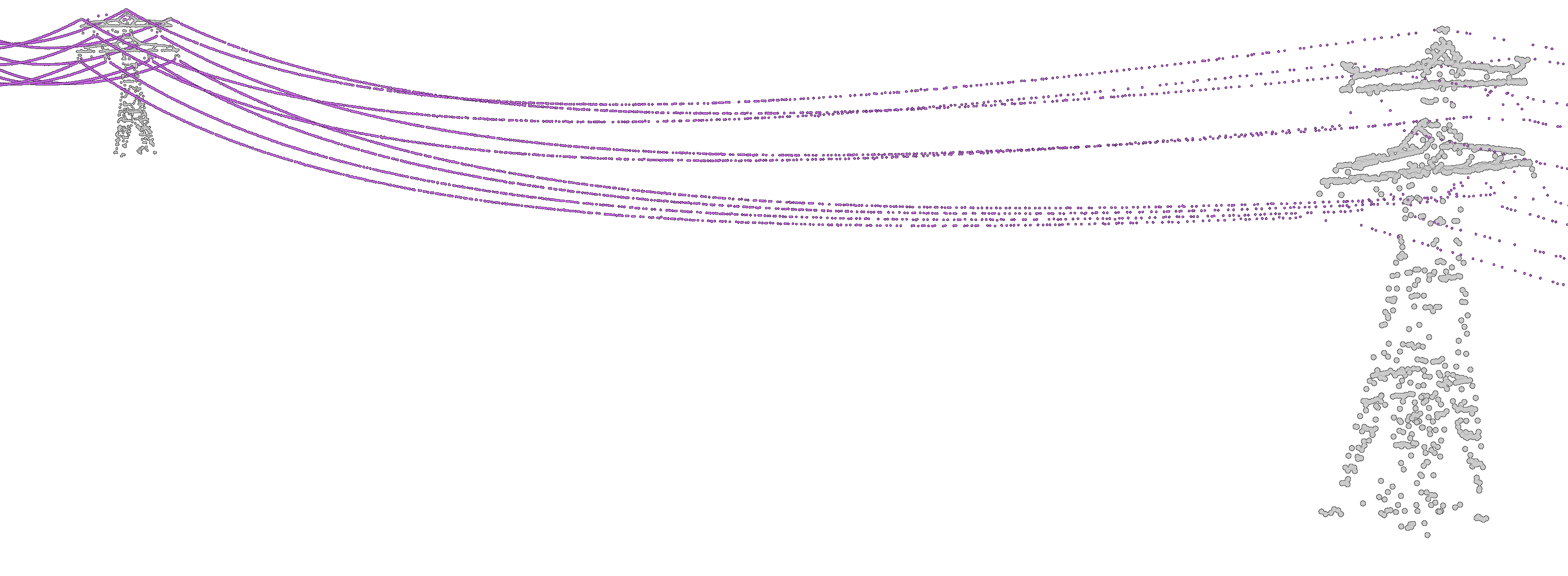}};
      \node at (0, 0) [above right, xshift = 0.5em, yshift = 0.5em] {\Large\textbf{a.}};
    \end{tikzpicture}
    &
    \begin{tikzpicture}
      \node at (0, 0) [above right] {\includegraphics[trim = {{300 px} 0 0 0}, clip, width = 0.4725\columnwidth, keepaspectratio]{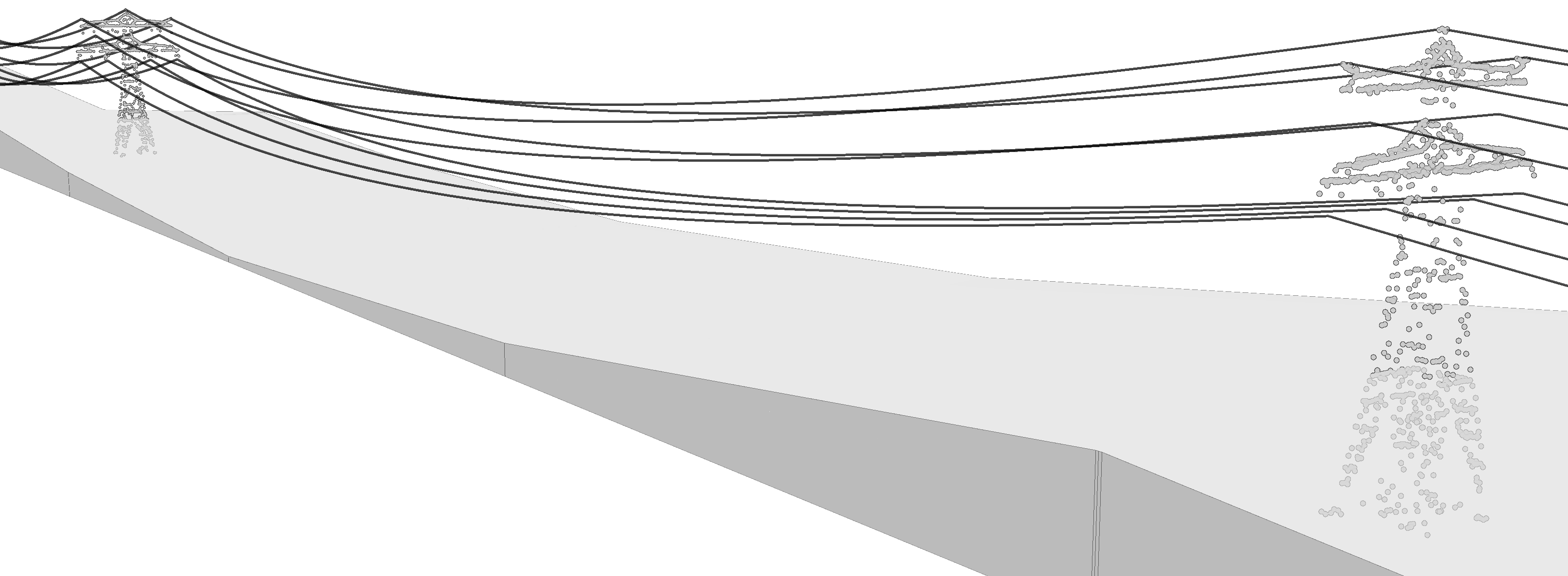}};
      \node at (0, 0) [above right, xshift = 0.5em, yshift = 0.5em] {\Large\textbf{b.}};
    \end{tikzpicture}
    \\
    \\
    \begin{tikzpicture}
      \node at (0, 0) [above right] {\includegraphics[trim = {{300 px} 0 0 0}, clip, width = 0.4725\columnwidth, keepaspectratio]{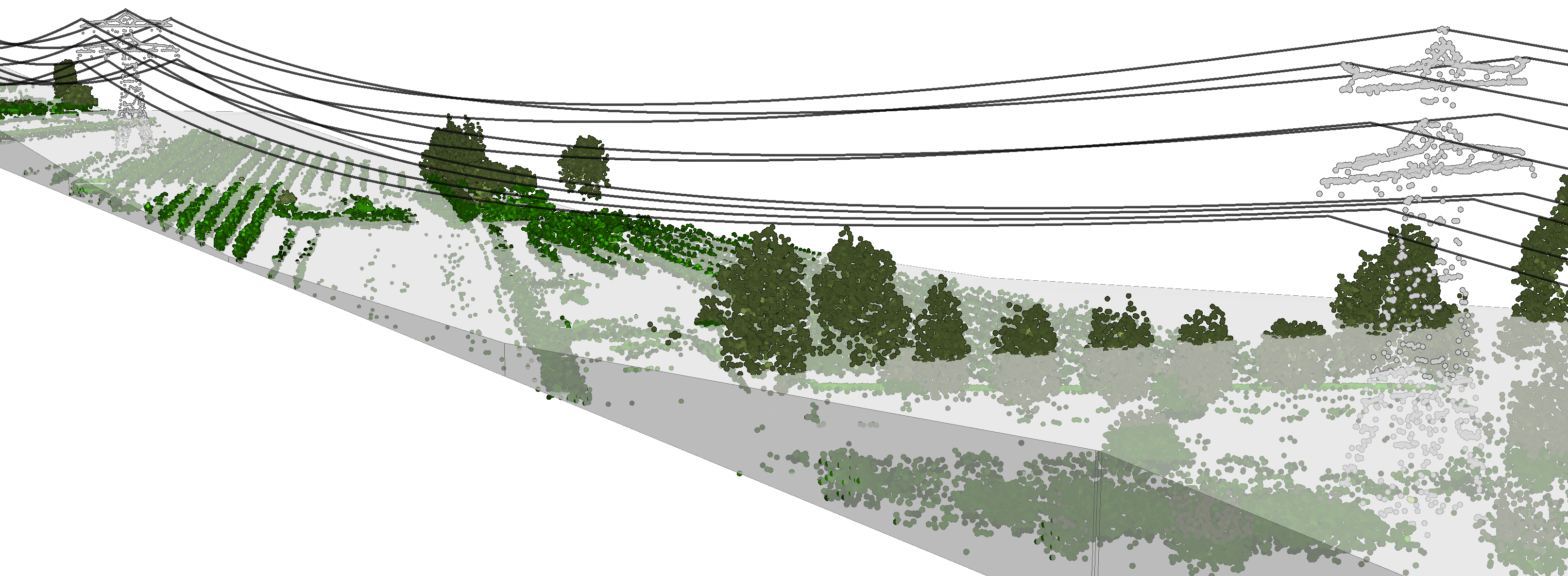}};
      \node at (0, 0) [above right, xshift = 0.5em, yshift = 0.5em] {\Large\textbf{c.}};
    \end{tikzpicture}
    &
    \begin{tikzpicture}
      \node at (0, 0) [above right] {\includegraphics[trim = {{300 px} 0 0 0}, clip, width = 0.4725\columnwidth, keepaspectratio]{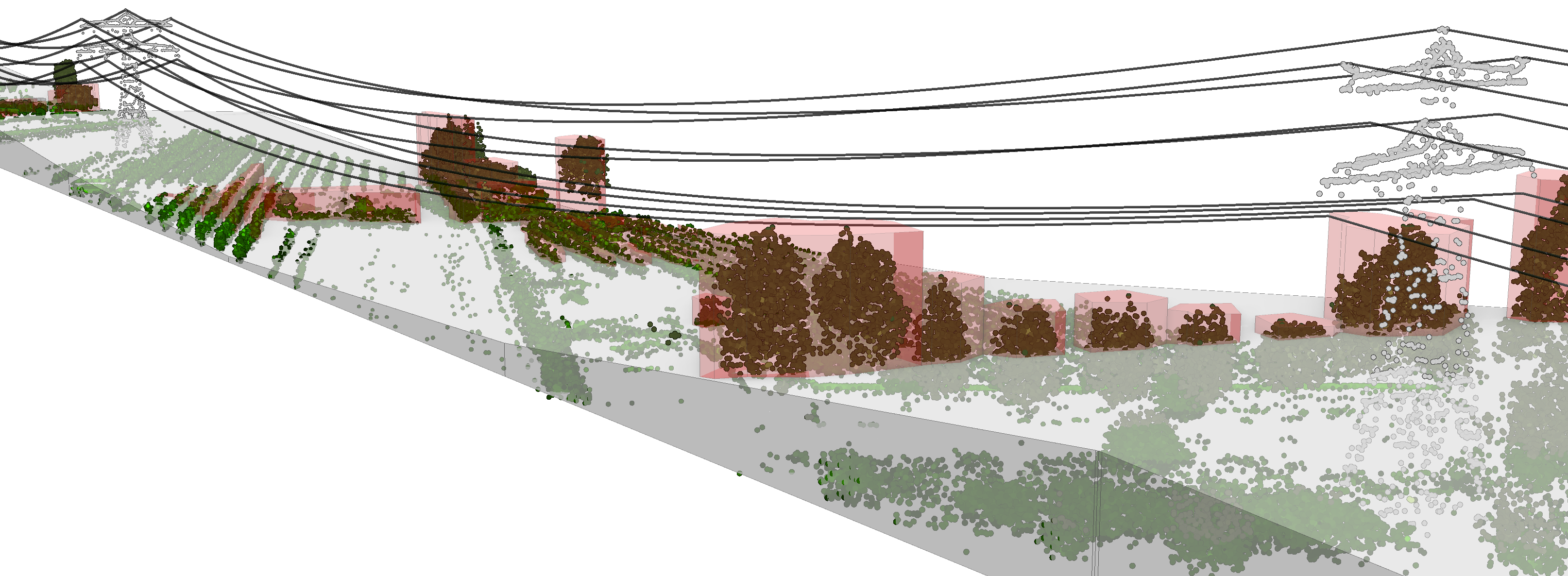}};
      \node at (0, 0) [above right, xshift = 0.5em, yshift = 0.5em] {\Large\textbf{d.}};
    \end{tikzpicture}
  \end{tabular}
  \captionsetup{singlelinecheck=off}
  \caption[caption]
  {
    Screenshots of results obtained in ArcGIS Pro when applying the catenary curve modeling solution to a workflow for managing vegetation growth around power lines.
    \begin{enumerate} [label = \alph*., ref = \alph*]
      \item Lidar points classified as power lines are drawn in purple, and transmission towers are drawn in gray.
      \item This screenshot shows line features constructed from the power line classified points and the clearance surface constructed from the line features.
      \item Vegetation points are displayed against a semitransparent clearance surface. Vegetation points within the clearance surface are tinted gray.
      \item Vegetation points exceeding the clearance surface are enclosed within a red bounding box.
    \end{enumerate}
  }
  \label{fig:PowerLinesProcessing}
\end{sidewaysfigure}

This solution can be demonstrated using a lidar survey collected over Utrecht, the Netherlands~\cite{ReferenceCurrentHeightFileNetherlands}, which captures a series of transmission power lines. The points representing power lines were labeled using a PointCNN classification model; see Figure~\ref{fig:PowerLinesProcessing}a. Line features modeling each power line's catenary curve were obtained by using the Extract Power Lines From Point Cloud tool in ArcGIS~Pro~\cite{ReferenceExtractPowerLinesFromPointCloud}, which implemented the solution discussed in this paper.
 
\begin{enumerate} [label = \arabic*., ref = \arabic*]
  \item The Extract Power Lines From Point Cloud tool was executed with the following parameters:
  \begin{itemize}
    \item Power line class code: $ 14 $
    \item Point tolerance: $ \SI{80}{\centi\metre} $
    \item Wire separation distance: $ \SI{1}{\metre} $
    \item Maximum wire sampling gap: $ \SI{15}{\metre} $
    \item Output line tolerance: $ \SI{1}{\centi\metre} $
    \item Minimum span for wind correction: $ \SI{60}{\metre} $
    \item Maximum deviation angle: $ 10\degree $
    \item End point search radius: $ \SI{10}{\metre} $
    \item Minimum wire length: $ \SI{5}{\metre} $
  \end{itemize}
  \item The tool generated a unique catenary curve for each power line; see Figure~\ref{fig:PowerLinesProcessing}b.
  \item The resultant power line was used to create a surface modeling a $ 15 $-meter horizontal clearance from the outermost power lines and a $ 9 $-meter vertical clearance from the lowest power lines; see Figure~\ref{fig:PowerLinesProcessing}c.
  \item This clearance surface was then used alongside a surface model of vegetation cover to evaluate the volume and area of the trees that exceed the clearance surface; see Figure~\ref{fig:PowerLinesProcessing}d. These results provide a basis for determining the quantity of vegetation that has exceeded the clearance zone.
\end{enumerate}

The aforementioned example demonstrates how the solution presented in this paper can be used to extract power lines as part of a downstream application for  evaluating vegetation encroachment.

\section{Conclusion}

This paper describes a solution for accurately extracting unique catenary curves from an array of points representing multiple catenary curves. The solution can be used to extract power lines from lidar surveys to support downstream applications. An example illustrating the potential use of this solution was provided in the examination of vegetation encroachment around transmission power lines.

Also in this paper, a new, efficient algorithm to find the closest point on the catenary curve is introduced, along with a~new algorithm for fitting a catenary curve to a set of points based on the shortest distance. This algorithm fits a catenary curve to points by approximating the shortest distance as a scalar product of a normal and the vector; see Section~\ref{sec:FittingCatenaryCurveToPoints}. Such an approach provides a general framework that can be applied to fitting other curves. The clustering algorithm is based on a minimum spanning tree, a dynamic programming approach, and the $ k $\=/mean clustering technique.

\section*{Acknowledgments}

The authors would like to thank Lois Stuart and Linda Thomas for proofreading this paper; and Manoj Lnu, senior software engineer, 3D analyst at Esri, for helpful discussions about the initial fitting of a catenary curve to a set of points.

\newcounter{CurrentSectionValue}
\setcounter{CurrentSectionValue}{\value{section}}
\setcounter{section}{0}
\renewcommand{\thesection}{Appendix \Roman{section}}

\section{Equation \eqref{equation:OYcrossing}\label{appendix:equation:crossing}}

\begin{equation*}
  y = \cosh{\left( x \right)} + \dfrac{x}{\sinh{\left( x \right)}}
  ,
\end{equation*}

\begin{equation}
  \begin{aligned}
    y' = & \sinh{\left( x \right)} + \dfrac{\sinh{\left( x \right)} - x \cosh{\left( x \right)}}{\sinh^2{\left( x \right)}} =\\
    = & \dfrac{\sinh^3{\left( x \right)} + \sinh{\left( x \right)} - x \cosh{\left( x \right)}}{\sinh^2{\left( x \right)}} =\\
    = & \dfrac{\sinh{\left( x \right)} \left( \sinh^2{\left( x \right)} + 1 \right) - x \cosh{\left( x \right)}}{\sinh^2{\left( x \right)}} =\\
    = & \dfrac{\sinh{\left( x \right)} \cosh^2{\left( x \right)} - x \cosh{\left( x \right)}}{\sinh^2{\left( x \right)}} =\\
    = & \dfrac{\cosh{\left( x \right)}}{\sinh^2{\left( x \right)}} \left( \sinh{\left( x \right)} \cosh{\left( x \right)} - x \right) =\\
    = & \dfrac{\cosh{\left( x \right)}}{\sinh^2{\left( x \right)}} \left( \dfrac{\sinh{\left( 2x \right)}}{2} - x \right)
    .
  \end{aligned}
  \label{equation:OYcrossing1}
\end{equation}

Because $ \sinh{\left( x \right)} - x $ is a strictly increasing odd function, it follows that
\eqref{equation:OYcrossing1} is positive for all positive values and negative for all negative values; therefore, \eqref{equation:OYcrossing} is strictly increasing for all positive values and strictly decreasing for all negative values.

The minimum point of the intersection of the $ y $\=/axis is always above $ 2 $, because
\begin{equation*}
  \lim_{x \to 0}{\left( \cosh{\left( x \right)} + \dfrac{x}{\sinh{\left( x \right)}} \right)} = 2
  .
\end{equation*}

\section{Length along a Parabola\label{appendix:length_along_parabola}}

For the parabola
\begin{equation}
  y = \dfrac{1}{2} x^2
  \label{eq:parabola_1_2}
\end{equation}
the length from point $ \myvectordef{ 0 }{ 0 } $ to point $ \myvectordef{ x }{ \dfrac{1}{2} x^2 } $ is calculated as the next integral
\begin{equation*}
  \operatorname{L}{\left( x \right)}
  =
  \int\limits_{0}^{x}
  {
    \sqrt{1 + t^2} \cdot dt
  }
  .
\end{equation*}

The solution for this integral can be found in \cite[p.~225, eq.~42]{MathematicalFormulas}.
\begin{equation*}
  \int
  {
    \sqrt{x^2 + a^2} \cdot dx
    =
    \dfrac{x}{2}
    \sqrt{x^2 + a^2}
    +
    \dfrac{a^2}{2}
    \log{\left| x + \sqrt{x^2 + a^2} \right|}
  }
  +
  C
  ,
\end{equation*}
where $ C $ is an arbitrary constant.

Taking $ a = 1 $ and replacing $ x $ with an absolute value of $ x $ to stabilize the evaluation of this integral for negative values (see the note on the evaluation of $ \arcsinh $ in \cite[chapter~5.6]{NumericalRecipes}), we obtain
\begin{equation}
  \operatorname{L}{\left( x \right)}
  =
  \sign{\left( x \right)} \cdot \dfrac{\left| x \right| \cdot \sqrt{x^2 + 1} + \log{\left( \left| x \right| + \sqrt{x^2 + 1} \right)}}{2}
  .
  \label{eq:parabola_length_integral_stable}
\end{equation}

Therefore, the length along any parabola
\begin{equation*}
  y = a \cdot x^2 + b \cdot x + c
\end{equation*}
between $ x_0 $ and $ x_1 $ can be found by shifting and scaling to match the parabola defined in \eqref{eq:parabola_1_2}. Using \eqref{eq:parabola_length_integral_stable}, we obtain
\begin{equation*}
  \operatorname{L}{\left( x_0, x_1, a, b \right)}
  =
  \dfrac
  {
    L{\left( 2 \cdot a \cdot x_1 + b \right)}
    -
    L{\left( 2 \cdot a \cdot x_0 + b \right)}
  }
  {
    2 \cdot a
  }
  .
\end{equation*}

\section{On the Roots of the Cubic Equation \eqref{eq:CubicEquationForOsculatingParabola}\label{RootsForCubicEquationForOsculatingParabola}}

\begin{figure*} [htb]
  \centering
  \begin{tikzpicture} [scale = 1.5]
    \tkzInit[xmin = -4.25, ymin = -0.25, xmax = 4.25, ymax = 5.25]

    \tkzGrid

    \begin{scope}
      \clip (-4.25, -0.25) rectangle (4.25, 5.25);
      \input{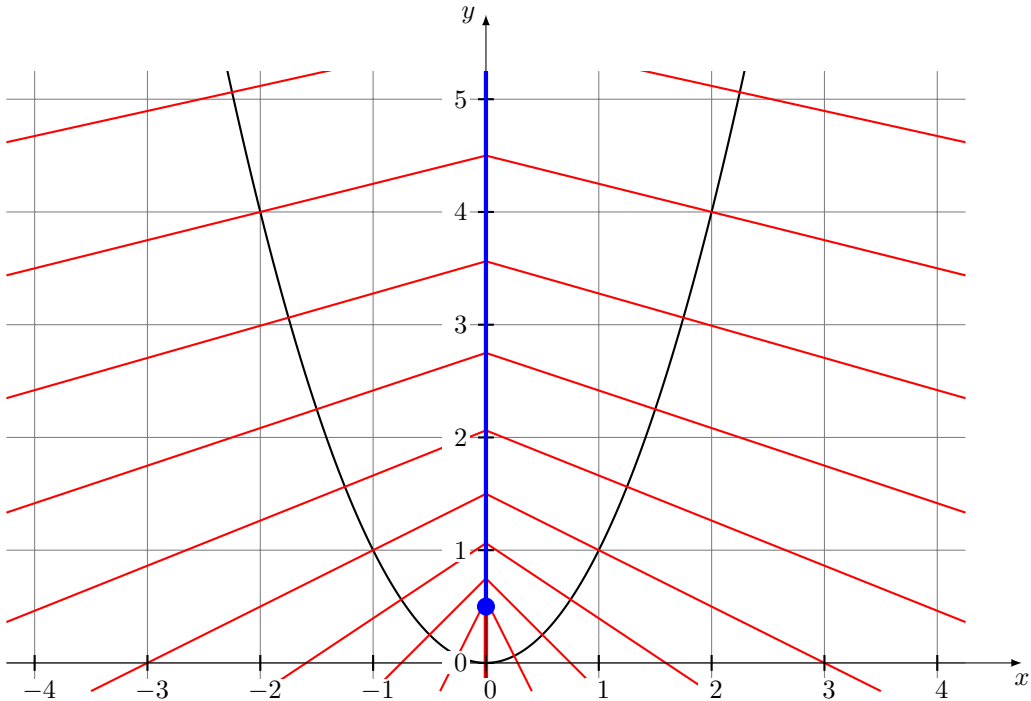}
        \draw [ultra thick, red] (-0.0, -0.25) -- (0.0, 0.5) -- (0.0, -0.25);
  \draw [thick, red] (-0.40625, -0.25) -- (0.0, 0.5625) -- (0.40625, -0.25);
  \draw [thick, red] (-1.0, -0.25) -- (0.0, 0.75) -- (1.0, -0.25);
  \draw [thick, red] (-1.96875, -0.25) -- (0.0, 1.0625) -- (1.96875, -0.25);
  \draw [thick, red] (-3.5, -0.25) -- (0.0, 1.5) -- (3.5, -0.25);
  \draw [thick, red] (-4.25, 0.36250000000000004) -- (0.0, 2.0625) -- (4.25, 0.36250000000000004);
  \draw [thick, red] (-4.25, 1.3333333333333333) -- (0.0, 2.75) -- (4.25, 1.3333333333333333);
  \draw [thick, red] (-4.25, 2.3482142857142856) -- (0.0, 3.5625) -- (4.25, 2.3482142857142856);
  \draw [thick, red] (-4.25, 3.4375) -- (0.0, 4.5) -- (4.25, 3.4375);
  \draw [thick, red] (-4.25, 4.618055555555555) -- (0.0, 5.5625) -- (4.25, 4.618055555555555);
  \draw [thick, red] (-4.25, 5.9) -- (0.0, 6.75) -- (4.25, 5.9);
  \draw [thick, red] (-4.25, 7.2897727272727275) -- (0.0, 8.0625) -- (4.25, 7.2897727272727275);
  \draw [thick, red] (-4.25, 8.791666666666666) -- (0.0, 9.5) -- (4.25, 8.791666666666666);
  \draw [thick, red] (-4.25, 10.408653846153847) -- (0.0, 11.0625) -- (4.25, 10.408653846153847);
  \draw [thick, red] (-4.25, 12.142857142857142) -- (0.0, 12.75) -- (4.25, 12.142857142857142);
  \draw [thick, red] (-4.25, 13.995833333333334) -- (0.0, 14.5625) -- (4.25, 13.995833333333334);
  \draw [thick, red] (-4.25, 15.96875) -- (0.0, 16.5) -- (4.25, 15.96875);
  \draw [thick, red] (-4.25, 18.0625) -- (0.0, 18.5625) -- (4.25, 18.0625);
  \draw [thick, red] (-4.25, 20.27777777777778) -- (0.0, 20.75) -- (4.25, 20.27777777777778);
  \draw [thick, red] (-4.25, 22.61513157894737) -- (0.0, 23.0625) -- (4.25, 22.61513157894737);
  \draw [thick, red] (-4.25, 25.075) -- (0.0, 25.5) -- (4.25, 25.075);
    \end{scope}

    \tkzAxeXY

    \begin{scope}
      \clip (-4.25, -0.25) rectangle (4.25, 5.25);
      \draw [ultra thick, blue] (0, 0.5) -- (0, 100);
      \draw [blue, fill = blue] (0, 0.5) circle (0.075);
    \end{scope}
  \end{tikzpicture}
  \caption
  {
    The relationship between the points and the closest points on the parabola ($ y = x^2 $).
    The~black line is a~parabola; the red lines are normal lines to the parabola; the blue line is where two points on the catenary curve have the same distance; and the blue dot is an area of instability for finding the closest point on the parabola.
  }
  \label{fig:NormalsToParabola}
\end{figure*}

Figure~\ref{fig:NormalsToParabola}, like Figure~\ref{fig:NormalsToCatenaryCurve}, shows the relationship between the points and the closest position on the parabola. This~relationship holds for different parabola shapes.

Let's define a parabola as
\begin{equation*}
  y = a \cdot x^2
  ,
\end{equation*}
where $ a > 0 $ is a coefficient.
The normal line will intersect the $ y $\=/axis at
\begin{equation*}
  y = a \cdot x^2 + \dfrac{1}{2 \cdot a}
  .
\end{equation*}

From this, it follows that the normal lines constructed for the points on the parabola for half-plane $ x > 0 $ do not intersect in this half-plane. It is the same as for the catenary curve. Therefore, there is only one closest point for points in the half-plane. Also, it follows that if there are three real roots for the cubic equation for points in the half-plane, two of them are negative and one is positive. For our task, we should only consider finding the closest point on the parabola's right branch because the left side of the parabola is not a good approximation of the catenary curve. Therefore, when looking for the closest point on the parabola and the point is located in another half-plane $ x \leq 0 $, there might be several positive roots, but the one with the largest value will be used.

\setcounter{section}{\value{CurrentSectionValue}}
\renewcommand{\thesection}{\Roman{section}}

\bibliographystyle{IEEEtran}
\bibliography{CatenaryCurve}

\begin{thebibliography}{10}
\providecommand{\url}[1]{#1}
\csname url@samestyle\endcsname
\providecommand{\newblock}{\relax}
\providecommand{\bibinfo}[2]{#2}
\providecommand{\BIBentrySTDinterwordspacing}{\spaceskip=0pt\relax}
\providecommand{\BIBentryALTinterwordstretchfactor}{4}
\providecommand{\BIBentryALTinterwordspacing}{\spaceskip=\fontdimen2\font plus
\BIBentryALTinterwordstretchfactor\fontdimen3\font minus
  \fontdimen4\font\relax}
\providecommand{\BIBforeignlanguage}[2]{{%
\expandafter\ifx\csname l@#1\endcsname\relax
\typeout{** WARNING: IEEEtran.bst: No hyphenation pattern has been}%
\typeout{** loaded for the language `#1'. Using the pattern for}%
\typeout{** the default language instead.}%
\else
\language=\csname l@#1\endcsname
\fi
#2}}
\providecommand{\BIBdecl}{\relax}
\BIBdecl

\bibitem{ReferenceCurrentHeightFileNetherlands}
\BIBentryALTinterwordspacing
\BIBforeignlanguage{dutch}{{AHN3} downloads. {A}ctueel {H}oogtebestand
  {N}ederland (current height file, {N}etherlands)}. [Online]. Available:
  \url{https://app.pdok.nl/ahn3-downloadpage/}
\BIBentrySTDinterwordspacing

\bibitem{ReferenceCalculationOfSag}
\BIBentryALTinterwordspacing
A.~Hatibovic, ``\BIBforeignlanguage{english}{Derivation of equations for
  conductor and sag curves of an overhead line based on a given catenary
  constant},'' \emph{\BIBforeignlanguage{english}{Periodica Polytechnica
  Electrical Engineering and Computer Science}}, vol.~58, no.~1, pp. 23--27,
  2014. [Online]. Available: \url{https://doi.org/10.3311/PPee.6993}
\BIBentrySTDinterwordspacing

\bibitem{ReferenceWildfireActivityStatistics2020}
\BIBentryALTinterwordspacing
\BIBforeignlanguage{english}{2020 {W}ildfire {A}ctivity {S}tatistics}.
  California Department of Forestry and Fire Protection. [Online]. Available:
  \url{https://www.fire.ca.gov/media/0fdfj2h1/2020_redbook_final.pdf}
\BIBentrySTDinterwordspacing

\bibitem{ReferenceWildfireActivityStatistics2019}
\BIBentryALTinterwordspacing
\BIBforeignlanguage{english}{2019 {W}ildfire {A}ctivity {S}tatistics}.
  California Department of Forestry and Fire Protection. [Online]. Available:
  \url{https://www.fire.ca.gov/media/iy1gpp2s/2019_redbook_final.pdf}
\BIBentrySTDinterwordspacing

\bibitem{ReferenceWildfireActivityStatistics2018}
\BIBentryALTinterwordspacing
\BIBforeignlanguage{english}{2018 {W}ildfire {A}ctivity {S}tatistics}.
  California Department of Forestry and Fire Protection. [Online]. Available:
  \url{https://www.fire.ca.gov/media/11146/2018_redbook_final.pdf}
\BIBentrySTDinterwordspacing

\bibitem{ReferenceWildfireActivityStatistics2017}
\BIBentryALTinterwordspacing
\BIBforeignlanguage{english}{2017 {W}ildfire {A}ctivity {S}tatistics}.
  California Department of Forestry and Fire Protection. [Online]. Available:
  \url{https://www.fire.ca.gov/media/10059/2017_redbook_final.pdf}
\BIBentrySTDinterwordspacing

\bibitem{ReferenceBoruvka1}
\BIBentryALTinterwordspacing
{O}takar {B}orůvka, ``O jistém problému minimálním,'' \emph{Práce
  Moravské přírodovědecké společnosti}, vol. III, no.~3, pp. 37--58,
  1926, {C}zech, {G}erman summary. [Online]. Available:
  \url{http://hdl.handle.net/10338.dmlcz/500114}
\BIBentrySTDinterwordspacing

\bibitem{ReferenceBoruvka2}
\BIBentryALTinterwordspacing
------, ``Příspěvek k otázce ekonomické stavby elektrovodných sítí,''
  \emph{Elektrotechnický obzor}, vol.~15, pp. 153--154, 1926, {C}zech,
  {G}erman summary. [Online]. Available:
  \url{http://hdl.handle.net/10338.dmlcz/500188}
\BIBentrySTDinterwordspacing

\bibitem{ReferenceOnTheShortestSpanningSubtreeOfAGraphAndTheTravelingSalesmanProblem}
\BIBentryALTinterwordspacing
J.~B. Kruskal, ``On the shortest spanning subtree of a graph and the traveling
  salesman problem,'' \emph{Proceedings of the American Mathematical Society},
  vol.~7, no.~1, pp. 48--50, 1956. [Online]. Available:
  \url{https://doi.org/10.1090/S0002-9939-1956-0078686-7}
\BIBentrySTDinterwordspacing

\bibitem{ReferenceShortestConnectionNetworksAndSomeGeneralizations}
\BIBentryALTinterwordspacing
R.~C. Prim, ``Shortest connection networks and some generalizations,''
  \emph{The Bell System Technical Journal}, vol.~36, no.~6, pp. 1389--1401,
  1957. [Online]. Available:
  \url{https://doi.org/10.1002/j.1538-7305.1957.tb01515.x}
\BIBentrySTDinterwordspacing

\bibitem{ReferenceBoruvka}
\BIBentryALTinterwordspacing
J.~Nešetřil, E.~Milková, and H.~Nešetřilová, ``{O}takar {B}orůvka on
  minimum spanning tree problem. {T}ranslation of both the 1926 papers,
  comments, history,'' \emph{Discrete Mathematics}, vol. 233, no. 1-3, pp.
  3--36, 2001, {C}zech and {S}lovak 2. [Online]. Available:
  \url{https://doi.org/10.1016/S0012-365X(00)00224-7}
\BIBentrySTDinterwordspacing

\bibitem{ReferenceKMean}
\BIBentryALTinterwordspacing
H.-H. Bock, ``\BIBforeignlanguage{english}{Origins and extensions of the $ k
  $-means algorithm in cluster analysis},''
  \emph{\BIBforeignlanguage{english}{Journal Electronique d'Histoire des
  Probabilit\'{e}s et de la Statistique / Electronic Journal for History of
  Probability and Statistics}}, vol.~4, no.~2, pp. 1--18, December 2008.
  [Online]. Available: \url{http://www.jehps.net/Decembre2008/Bock.pdf}
\BIBentrySTDinterwordspacing

\bibitem{PolylineGeneralizationCombinatorical}
\BIBentryALTinterwordspacing
A.~Gribov and E.~Bodansky, ``\BIBforeignlanguage{English}{A new method of
  polyline approximation},'' in \emph{\BIBforeignlanguage{English}{Structural,
  Syntactic, and Statistical Pattern Recognition}}, ser. Lecture Notes in
  Computer Science, A.~Fred, T.~M. Caelli, R.~P. Duin, A.~Campilho, and
  D.~de~Ridder, Eds.\hskip 1em plus 0.5em minus 0.4em\relax Springer Berlin
  Heidelberg, 2004, vol. 3138, pp. 504--511. [Online]. Available:
  \url{https://doi.org/10.1007/978-3-540-27868-9_54}
\BIBentrySTDinterwordspacing

\bibitem{NumericalRecipes}
\BIBentryALTinterwordspacing
W.~H. Press, S.~A. Teukolsky, W.~T. Vetterling, and B.~P. Flannery,
  \emph{\BIBforeignlanguage{english}{Numerical Recipes, Third Edition: The Art
  of Scientific Computing}}.\hskip 1em plus 0.5em minus 0.4em\relax Cambridge
  University Press, 2007. [Online]. Available:
  \url{www.cambridge.org/9780521880688}
\BIBentrySTDinterwordspacing

\bibitem{ReferenceBookTrustRegionMethods}
\BIBentryALTinterwordspacing
A.~R. Conn, N.~I.~M. Gould, and P.~L. Toint, \emph{Trust Region Methods}.\hskip
  1em plus 0.5em minus 0.4em\relax Society for Industrial and Applied
  Mathematics, 2000. [Online]. Available:
  \url{https://doi.org/10.1137/1.9780898719857}
\BIBentrySTDinterwordspacing

\bibitem{ReferenceLeastSquareFittingKennethLevenberg0}
\BIBentryALTinterwordspacing
K.~Levenberg, ``\BIBforeignlanguage{english}{A method for the solution of
  certain non-linear problems in least squares},''
  \emph{\BIBforeignlanguage{english}{Quarterly of Applied Mathematics}},
  vol.~2, pp. 164--168, 1944. [Online]. Available:
  \url{https://doi.org/10.1090/qam/10666}
\BIBentrySTDinterwordspacing

\bibitem{ReferenceLeastSquareFittingDonaldWMarquardt1}
\BIBentryALTinterwordspacing
D.~W. Marquardt, ``\BIBforeignlanguage{english}{An algorithm for least-squares
  estimation of nonlinear parameters},''
  \emph{\BIBforeignlanguage{english}{Journal of the Society for Industrial and
  Applied Mathematics}}, vol.~11, no.~2, pp. 431--441, 1963. [Online].
  Available: \url{https://doi.org/10.1137/0111030}
\BIBentrySTDinterwordspacing

\bibitem{ReferenceLeastSquareFittingAnanthRanganathan2}
\BIBentryALTinterwordspacing
A.~Ranganathan. (2004) \BIBforeignlanguage{english}{The {L}evenberg-{M}arquardt
  algorithm}. [Online]. Available:
  \url{http://www.ananth.in/Notes_files/lmtut.pdf}
\BIBentrySTDinterwordspacing

\bibitem{ReferenceLeastSquareFittingHenriPGavin3}
\BIBentryALTinterwordspacing
H.~P. Gavin. (2020) \BIBforeignlanguage{english}{The {L}evenberg-{M}arquardt
  method for nonlinear least squares curve-fitting problems}. [Online].
  Available: \url{https://people.duke.edu/~hpgavin/ce281/lm.pdf}
\BIBentrySTDinterwordspacing

\bibitem{ReferenceExtractPowerLinesFromPointCloud}
\BIBentryALTinterwordspacing
Extract power lines from point cloud (3{D} {A}nalyst). Retrieved October, 2021.
  [Online]. Available:
  \url{pro.arcgis.com/en/pro-app/latest/tool-reference/3d-analyst/extract-power-lines-from-point-cloud.htm}
\BIBentrySTDinterwordspacing

\bibitem{MathematicalFormulas}
{\cyrВ}.~{\cyrТ}. Воднев, {\cyrА}.~{\cyrФ}. Наумович, and
  {\cyrН}.~{\cyrФ}. Наумович,
  \emph{\BIBforeignlanguage{russian}{Основные
  математические формулы (Main mathematical formulas)}},
  {\cyrЮ}.~{\cyrС}. Богданова, Ed.\hskip 1em plus 0.5em minus
  0.4em\relax Минск, Вышэйшая Школа, 1980.

\end{thebibliography}


\end{document}